\documentclass[11pt,a4paper]{article}
\usepackage[hyperref]{acl2021}
\usepackage{times}
\usepackage{latexsym}

\usepackage{microtype}

\aclfinalcopy 
\def\aclpaperid{3802} 

\usepackage{amsmath}
\usepackage{blindtext}
\usepackage{booktabs}
\usepackage{graphicx}
\usepackage{inconsolata}
\usepackage{soul}
\usepackage{subcaption}
\DeclareMathOperator*{\argmax}{arg\,max}

\title{Mind Your Outliers! Investigating the Negative Impact of Outliers on \\ Active Learning for Visual Question Answering}

\author{
    Siddharth Karamcheti \ \ \   Ranjay Krishna \ \ \  Li Fei-Fei \ \ \  Christopher D. Manning  \\
    Department of Computer Science, Stanford University \\
    { \fontfamily{txtt}\selectfont \{skaramcheti, ranjaykrishna, feifeili, manning\}@cs.stanford.edu }
} 

\begin{document}
\maketitle

\begin{abstract}
Active learning promises to alleviate the massive data needs of supervised machine learning: it has successfully improved sample efficiency by an order of magnitude on traditional tasks like topic classification and object recognition. However, we uncover a striking contrast to this promise: across 5 models and 4 datasets on the task of visual question answering, a wide variety of active learning approaches fail to outperform random selection. To understand this discrepancy, we profile 8 active learning methods on a per-example basis, and identify the problem as \textit{collective outliers} -- groups of examples that active learning methods prefer to acquire but models fail to learn (e.g., questions that ask about text in images or require external knowledge). Through systematic ablation experiments and qualitative visualizations, we verify that collective outliers are a general phenomenon responsible for degrading pool-based active learning. Notably, we show that active learning sample efficiency increases significantly as the number of collective outliers in the active learning pool decreases. We conclude with a discussion and prescriptive recommendations for mitigating the effects of these outliers in future work.
\end{abstract}

\section{Introduction}
\label{sec:introduction}
\begin{figure}[t!]
    \centering
    \includegraphics[width=0.95\columnwidth]{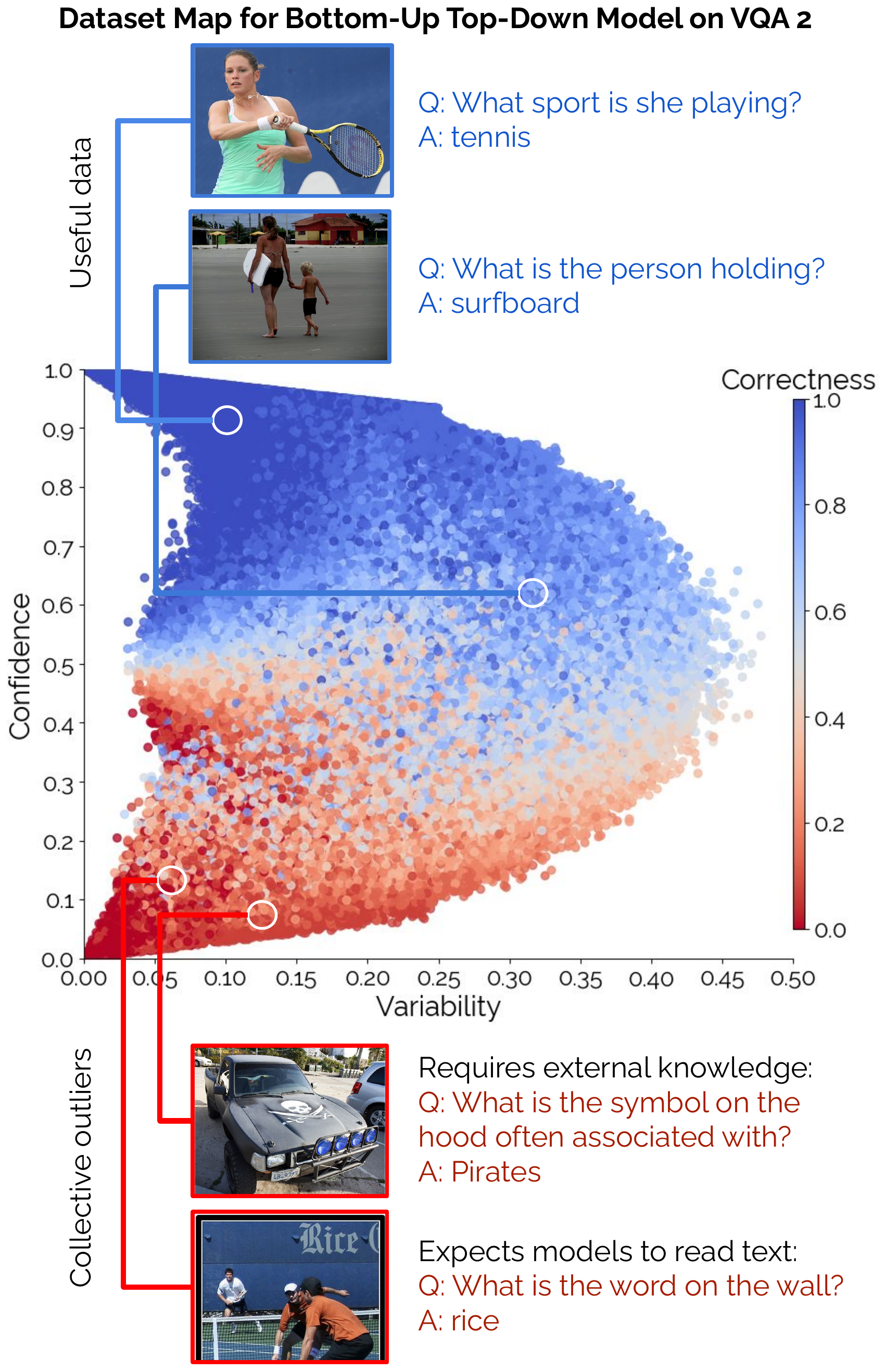}
    \caption{We systematically evaluate active learning on VQA datasets and isolate their inability to perform better than random sampling due to the presence of \textit{collective outliers}. Active learning methods prefer to acquire these outliers, which are hard and often impossible for models to learn. We show that Dataset Maps, like the one shown here, can heuristically identify these collective outliers as examples assigned low model confidence and prediction variability during training.}
    \label{fig:pull_figure}
\end{figure}

Today, language-equipped vision systems such as VizWiz, TapTapSee, BeMyEyes, and CamFind are actively being deployed across a broad spectrum of users.\footnote{
    Applications can be found at \url{https://vizwiz.org/}, \url{https://taptapsee.com/}, \url{https://www.bemyeyes.com/}, and \url{https://camfindapp.com/}
}
As underlying methods improve, these systems will be expected to operate over diverse visual environments and understand myriad language inputs \citep{bigham2010vizwiz,tellex2011understanding,mei2016listen,zhu2017target,anderson2018vision,park2019ai}. Visual Question Answering (VQA), the task of answering questions about visual inputs, is a popular benchmark used to evaluate progress towards such open-ended systems \citep{agrawal2015vqa1,krishna2017visual,gordon2018iqa,hudson2019gqa}. Unfortunately, today's VQA models are data hungry: Their performance scales monotonically with more training data \citep{lu2016hierarchical,lin2017active}, motivating the need for data acquisition mechanisms such as active learning, which maximize performance while minimizing expensive data labeling.

While active learning is often key to effective data acquisition when such labeled data is difficult to obtain \citep{lewis1994heterogeneous,tong2001support,culotta2005reducing,settles2009active}, we find that 8 modern active learning methods \citep{gal2017dbal,siddhant2018deep,lowell2019practical} show little to no improvement in sample efficiency across 5 models on 4 VQA datasets -- indeed, in some cases performing worse than randomly selecting data to label. This finding is in stark contrast to the successful application of active learning methods on a variety of traditional tasks, such as topic classification \citep{siddhant2018deep,lowell2019practical}, object recognition \citep{deng2018adversarial}, digit classification \citep{gal2017dbal}, and named entity recognition \citep{shen2017deep}. Our negative results hold even when accounting for common active learning ailments: cold starts, correlated sampling, and uncalibrated uncertainty. We mitigate the cold start challenge of needing a representative initial dataset by varying the size of the seed set in our experiments. We account for sampling correlated data within a given batch by including Core-Set selection \citep{sener2018active} in the set of active learning methods we evaluate. Finally, we use deep Bayesian active learning to calibrate model uncertainty to high-dimensional data \citep{houlsby2011bayesian,gal2016dropout,gal2017dbal}.

After concluding that negative results are consistent across all experimental conditions, we investigate active learning's ineffectiveness on VQA as a data problem and identify the existence of \textit{collective outliers} \citep{han2000datamining} as the source of the problem. Leveraging recent advances in model interpretability, we build \textit{Dataset Maps} \citep{swayamdipta2020dataset}, which distinguish between collective outliers and useful data that improve validation set performance (see Figure~\ref{fig:pull_figure}). While global outliers deviate from the rest of the data and are often a consequence of labeling error, collective outliers cluster together; they may not individually be identifiable as outliers but collectively deviate from other examples in the dataset. For instance, VQA-2 \citep{goyal2017making} is riddled with collections of hard questions that require external knowledge to answer (e.g.,~``What is the symbol on the hood often associated with\@?'') or that ask the model to read text in the images (e.g.,~``What is the word on the wall?''). Similarly, GQA \citep{hudson2019gqa} asks underspecified questions (e.g.,~``what is the person wearing\@?'' which can have multiple correct answers). Collective outliers are not specific to VQA, but can similarly be found in many open-ended tasks, including visual navigation \citep{anderson2018vision} (e.g.,~``Go to the grandfather clock'' requires identifying rare grandfather clocks), and open-domain question answering \citep{kwiatkowski2019natural}, amongst others. 

Using Dataset Maps, we profile active learning methods and show that they prefer acquiring collective outliers that models are unable to learn, explaining their poor improvements in sample efficiency relative to random sampling. Building on this, we use these maps to perform ablations where we identify and remove outliers iteratively from the active learning pool, observing correlated improvements in sample efficiency. This allows us to conclude that collective outliers are, indeed, responsible for the ineffectiveness of active learning for VQA. We end with prescriptive suggestions for future work in building active learning methods robust to these types of outliers.

\section{Related Work}
\label{sec:related-work}
Our work tests the utility of multiple recent active learning methods on the open-ended understanding task of VQA. We draw on the dataset analysis literature to identify collective outliers as the bottleneck hindering active learning methods in this setting.

\paragraph{Active Learning.}
Active learning strategies have been successfully applied to image recognition \citep{joshi2009multi, sener2018active}, information extraction \citep{scheffer2001active,finn2003active,jones2003active,culotta2005reducing}, named entity recognition \citep{hachey2005investigating,shen2017deep}, semantic parsing \citep{dong2018confidence}, and text categorization \citep{lewis1994sequential, hoi2006batch}. However, these same methods struggle to outperform a random baseline when applied to the task of VQA \citep{lin2017active,jedoui2019deep}.
To study this discrepancy, we systematically apply 8 diverse active learning methods to VQA, including methods that use model uncertainty \citep{abramson2004active,collins2008towards,joshi2009multi}, Bayesian uncertainty \citep{gal2016dropout,kendall2017uncertainties}, disagreement \citep{houlsby2011bayesian,gal2017dbal}, and Core-Set selection \citep{sener2018active}. 

\paragraph{Visual Question Answering.}
Progress on VQA has been heralded as a marker for progress on general open-ended understanding tasks, resulting in several benchmarks \citep{agrawal2015vqa1,malinowski2015ask,ren2015exploring,johnson2017clevr,goyal2017making,krishna2017visual,suhr2019nlvr2,hudson2019gqa} and models \citep{zhou2015simple,fukui2016multimodal,lu2016hierarchical,yang2016stacked,zhu2016visual7w,wu2016ask,anderson2018butd,tan2019lxmert,chen2020uniter}. To ensure that our negative results are not dataset or model-specific, we sample 4 datasets and 5 representative models, each utilizing unique visual and linguistic features and employing different inductive biases.

\paragraph{Interpreting and Analyzing Datasets.}
Given the prevalence of large datasets in modern machine learning, it is critical to assess dataset properties
to remove redundancies \citep{gururangan2018annotation,li2019repair} or biases \citep{torralba2011unbiased,khosla2012undoing,bolukbasi2016man}, both of which negatively impact sample efficiency. 
Prior work has used training dynamics to find examples which are frequently forgotten \citep{krymolowski2002distinguishing, toneva2019empirical} versus those that are easy to learn \citep{le2020adversarial}. This work suggests using two model-specific measures -- \textit{confidence} and \textit{prediction variance} -- as indicators of a training example's ``learnability'' \citep{chang2017active,swayamdipta2020dataset}. Dataset Maps \citep{swayamdipta2020dataset}, a recently introduced framework uses these two measures to profile datasets to find learnable examples. Unlike prior datasets analyzed by Dataset Maps that have a small number of global outliers as hard examples, we discover that VQA datasets contain copious amounts of collective outliers, which are difficult or even impossible for models to learn.

\section{Active Learning Experimental Setup}
\label{sec:active-learning}
\begin{table}[t]
\centering
\begin{tabular}{@{}lrr@{}}
\toprule
           & Pool Size      & \# Answers \\ \midrule
VQA-Sports & 5,411 [5k]      & 20         \\
VQA-Food   & 4,082 [4k]      & 20         \\
VQA-2      & 411,272 [400k] & 3130       \\
GQA        & 943,000 [900k] & 1842       \\ \bottomrule
\end{tabular}
\caption{We evaluate active learning on 4 VQA datasets. We display the total available training examples, effective pool sizes we use [in brackets], and the total number of possible answers for each dataset.}
\label{tab:datasets}
\end{table}

We adopt the standard pool-based active learning setup from prior work \citep{lewis1994sequential,settles2009active,gal2017dbal,lin2017active}, consisting of a model $\mathcal{M}$, initial seed set of labeled examples $(x_i, y_i) \in \mathcal{D}_{\text{seed}}$ used to initialize $\mathcal{M}$, an unlabeled pool of data $\mathcal{D}_{\text{pool}}$, and an acquisition function $\mathcal{A}(x, \mathcal{M})$. We run active learning over a series of acquisition iterations $T$ where at each iteration we acquire a batch of $B$ new examples per: $\hat{x} \in \mathcal{D}_{\text{pool}}$ to label per $\hat{x} = \argmax_{x \in \mathcal{D}_{\text{pool}}} \mathcal{A}(x, \mathcal{M})$.

Acquiring an example often refers to using an oracle or human expert to annotate a new example with a correct label. We follow prior work to simulate an oracle using existing datasets, forming $\mathcal{D}_{\text{seed}}$ from a fixed percentage of the full dataset, and using the remainder as $\mathcal{D}_{\text{pool}}$ \citep{gal2017dbal,lin2017active,siddhant2018deep}. We re-train $\mathcal{M}$ after each acquisition iteration. 

Prior work has noted the impact of seed set size on active learning performance \citep{lin2017active,misra2018learning,jedoui2019deep}. We run multiple active learning evaluations with varying seed set sizes (ranging from 5\% to 50\% of the full pool size). We keep the size of each acquisition batch $B$ to a constant 10\% of the overall pool size.

\subsection{Models}
Visual Question Answering (VQA) requires reasoning over two modalities: images and text. Most models use feature ``backbones'' (e.g., features from object recognition models pretrained on ImageNet, and pretrained word vectors for text). For image features we use grid-based features from ResNet-101 \citep{he2016resnet}, or object-based features from Faster R-CNN \citep{ren2015frcnn} fine-tuned on Visual Genome \citep{anderson2018butd}. We evaluate with a representative sample of existing VQA models, including the following:\footnote{
    Key implementation details can be found in the appendix. In the interest of full reproducibility and further work in active learning and VQA, we release our code and results here: \url{https://github.com/siddk/vqa-outliers}. 
}

\paragraph{LogReg} is a logistic regression model that uses either ResNet-101 or Faster R-CNN image features with mean-pooled GloVe question embeddings \citep{pennington2014glove}. Although these models are not as performant as the subsequent models, logistic regression has been effective on VQA \citep{suhr2019nlvr2}, and is pervasive in the active learning literature \citep{schein2007active,yang2018benchmark,mussmann2018accuracy}.

\paragraph{LSTM-CNN} is a standard model introduced with VQA-1 \citep{agrawal2015vqa1}. We use more performant ResNet-101 features instead of the original VGGNet features as our visual backbone. 

\paragraph{BUTD} (Bottom-Up Top-Down Attention) uses object-based features in tandem with attention over objects \citep{anderson2018butd}. \textsc{BUTD} won the 2017 VQA Challenge \citep{teney2018tips}, and has been a consistent baseline for recent work in VQA. 

\paragraph{LXMERT} is a large multi-modal transformer model that uses \textsc{BUTD}'s object features and contextualized BERT \citep{devlin2019bert} language features \citep{tan2019lxmert}. LXMERT is pretrained on a corpus of aligned image-and-textual data spanning MS COCO, Visual Genome, VQA-2, NLVR-2, and GQA \citep{lin2014microsoft,krishna2017visual,goyal2017making,suhr2019nlvr2,hudson2019gqa}, initializing a cross-modal representation space conducive to fine-tuning.\footnote{
    Results for LXMERT in \citet{tan2019lxmert} are reported \textit{after} pretraining on training and validation examples from the VQA datasets we use. While this is fair if the goal is optimizing for test performance, this exposure to training and validation examples leaks important information; to remedy this, we obtained a model checkpoint from the LXMERT authors trained \textit{without} VQA data. This is also why our LXMERT results are lower than the numbers reported in the original paper -- however, the general boost provided by cross-modal pretraining holds.
}

\begin{figure*}
    \centering
    \begin{subfigure}[b]{0.31\textwidth}
        \centering
        \includegraphics[width=\textwidth, height=0.2\textheight]{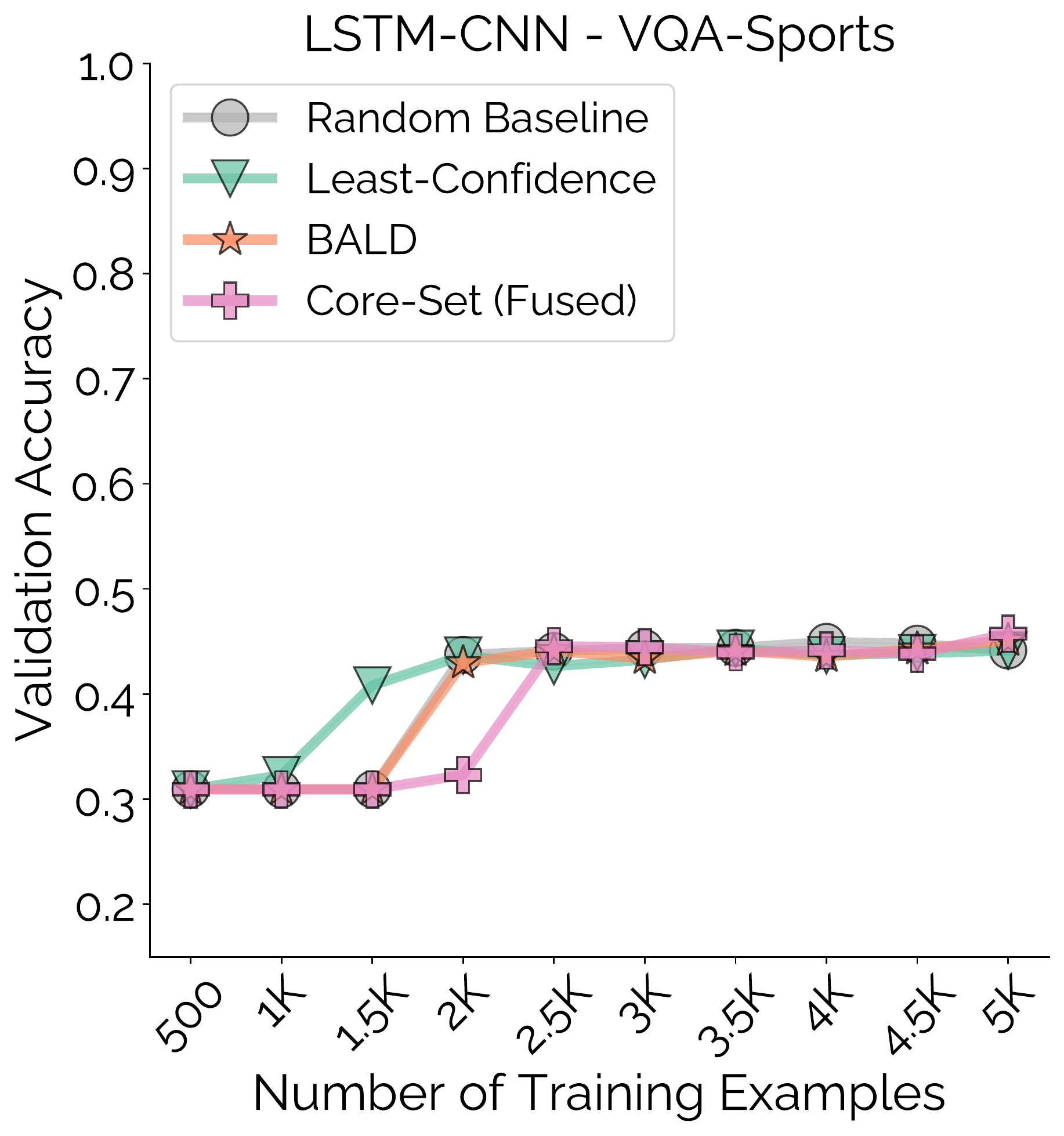}
    \end{subfigure}
    \hfill
    \begin{subfigure}[b]{0.31\textwidth}
        \centering
        \includegraphics[width=\textwidth, height=0.2\textheight]{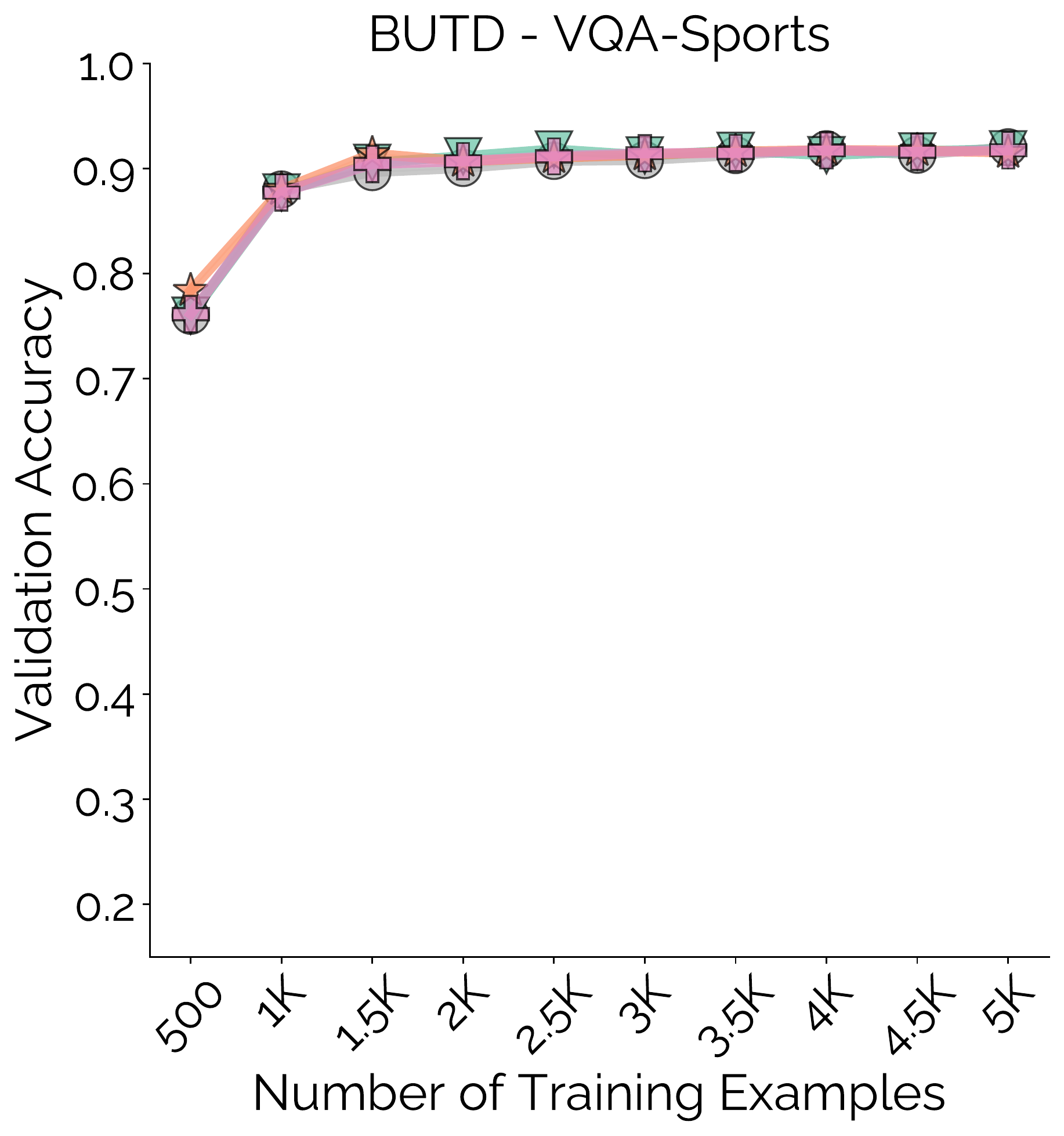}
    \end{subfigure}
    \hfill
    \begin{subfigure}[b]{0.31\textwidth}
        \centering
        \includegraphics[width=\textwidth, height=0.2\textheight]{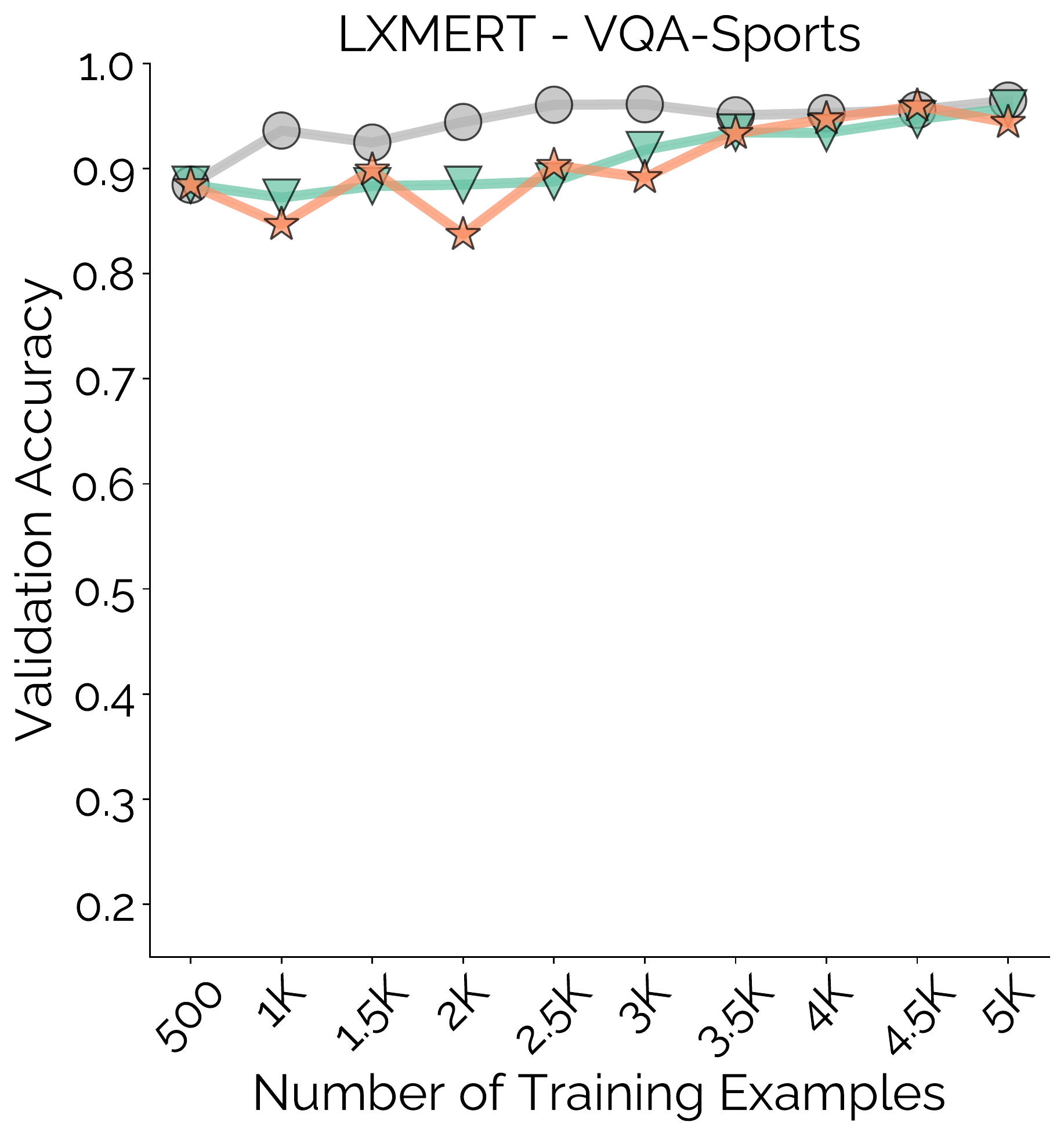}
    \end{subfigure}
    \vspace*{-5pt}
    \caption{Results for varied active learning methods on VQA-Sports, a simplified VQA dataset. Strategies perform on par with or worse than the random baseline, when using $10\%$ of the full dataset as the seed set.}
    \label{fig:vqa-sports-p10}
\end{figure*}

\begin{figure*}
    \centering
    \begin{subfigure}[b]{0.31\textwidth}
        \centering
        \includegraphics[width=\textwidth, height=0.2\textheight]{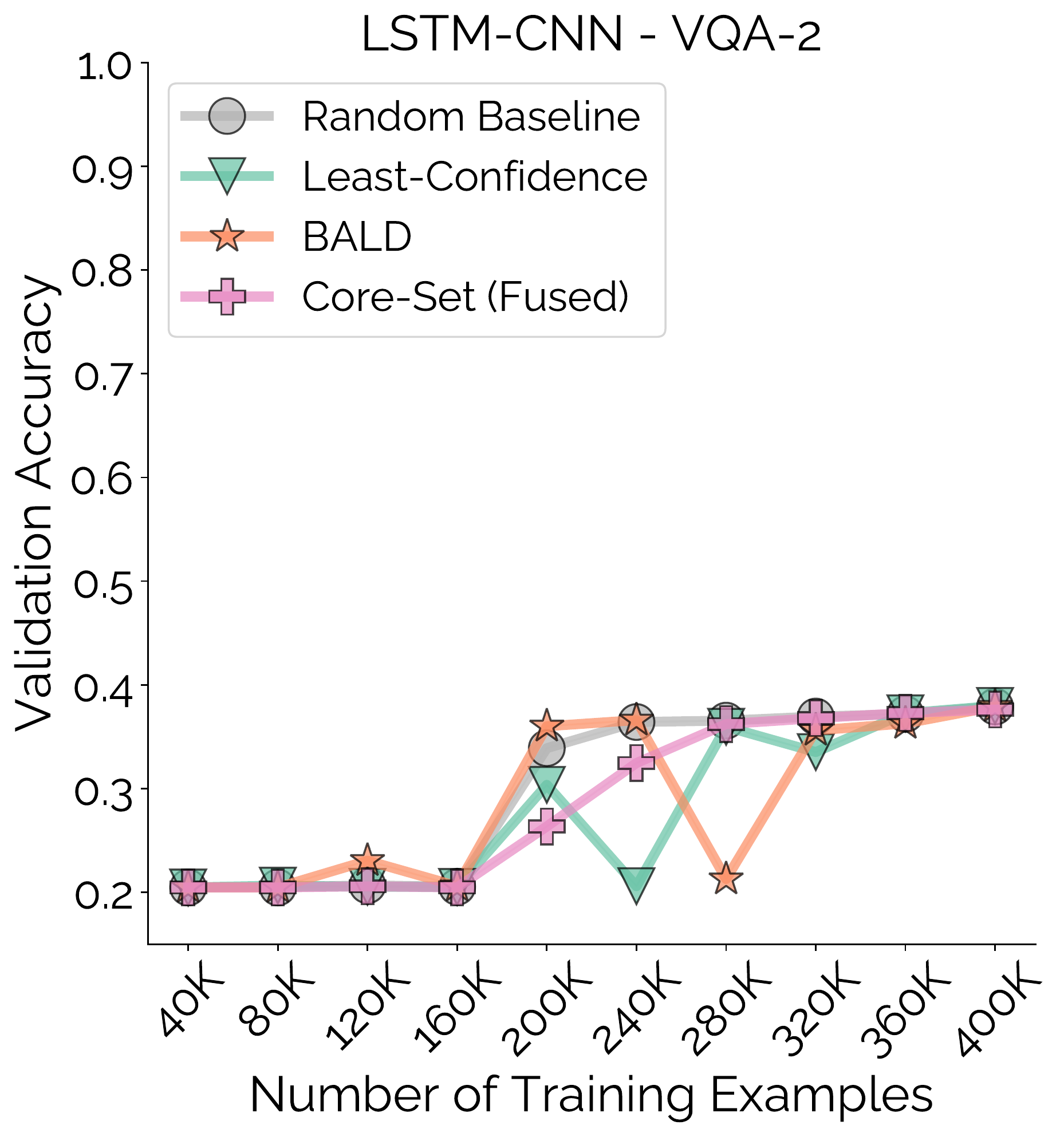}
    \end{subfigure}
    \hfill
    \begin{subfigure}[b]{0.31\textwidth}
        \centering
        \includegraphics[width=\textwidth, height=0.2\textheight]{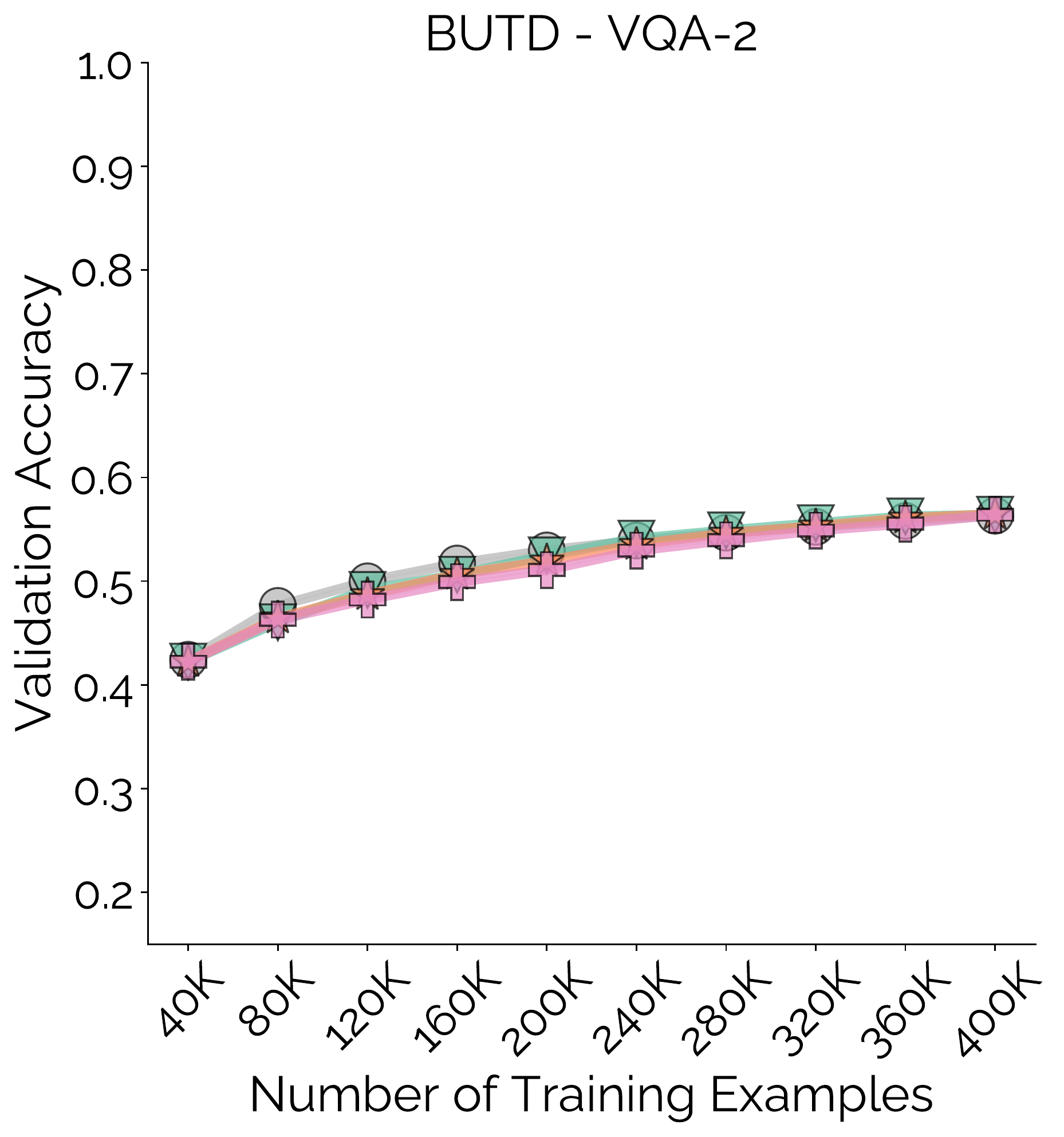}
    \end{subfigure}
    \hfill
    \begin{subfigure}[b]{0.31\textwidth}
        \centering
        \includegraphics[width=\textwidth, height=0.2\textheight]{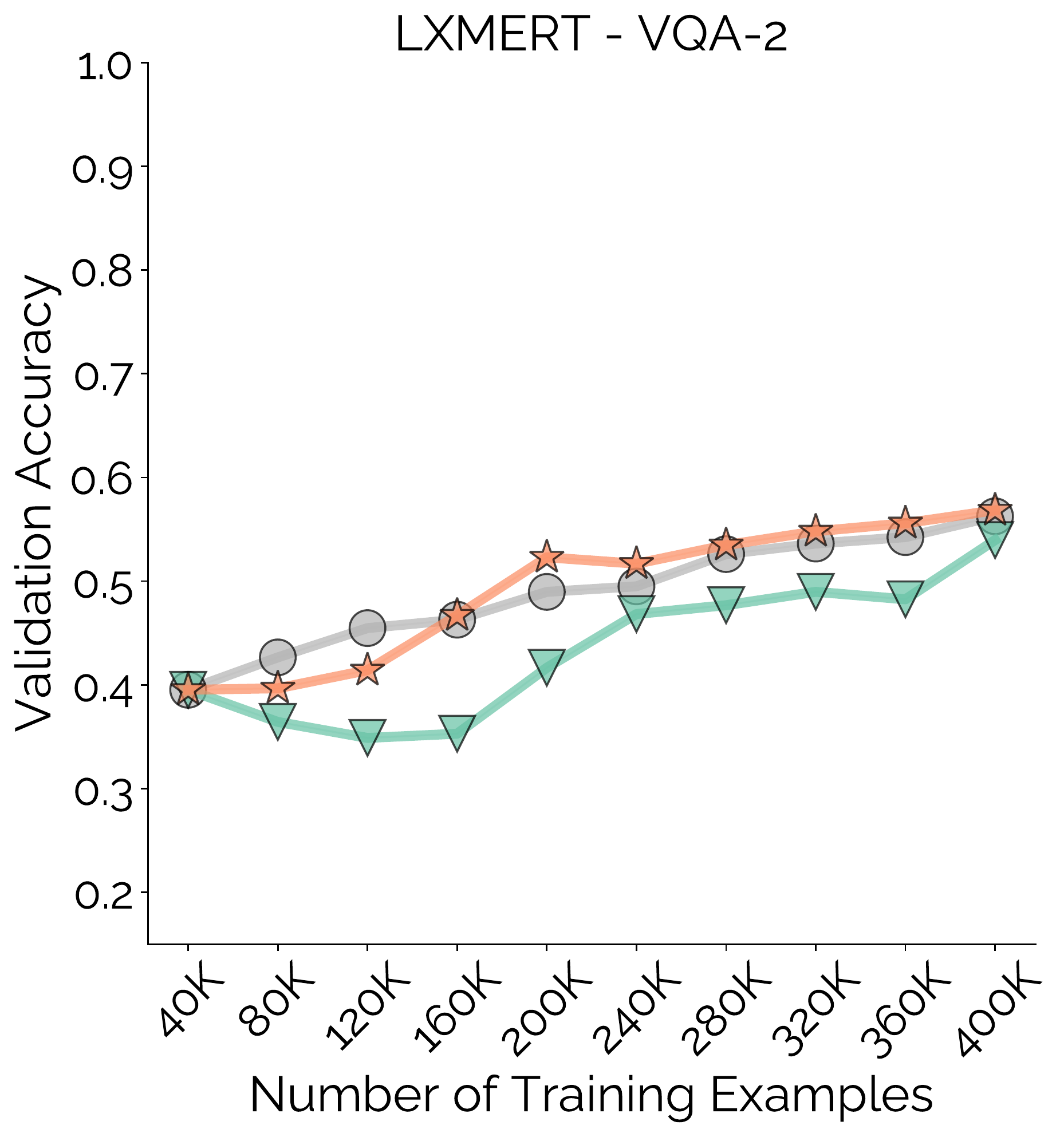}
    \end{subfigure}
    \vspace*{-5pt}
    \caption{Results for the full VQA-2 dataset, also using 10\% of the full dataset as a seed set. Similar to the plot above, all active learning methods perform similar to a random baseline.}    
    \label{fig:vqa2-p10}
\end{figure*}

\begin{figure*}
    \centering
    \begin{subfigure}[b]{0.31\textwidth}
        \centering
        \includegraphics[width=\textwidth, height=0.2\textheight]{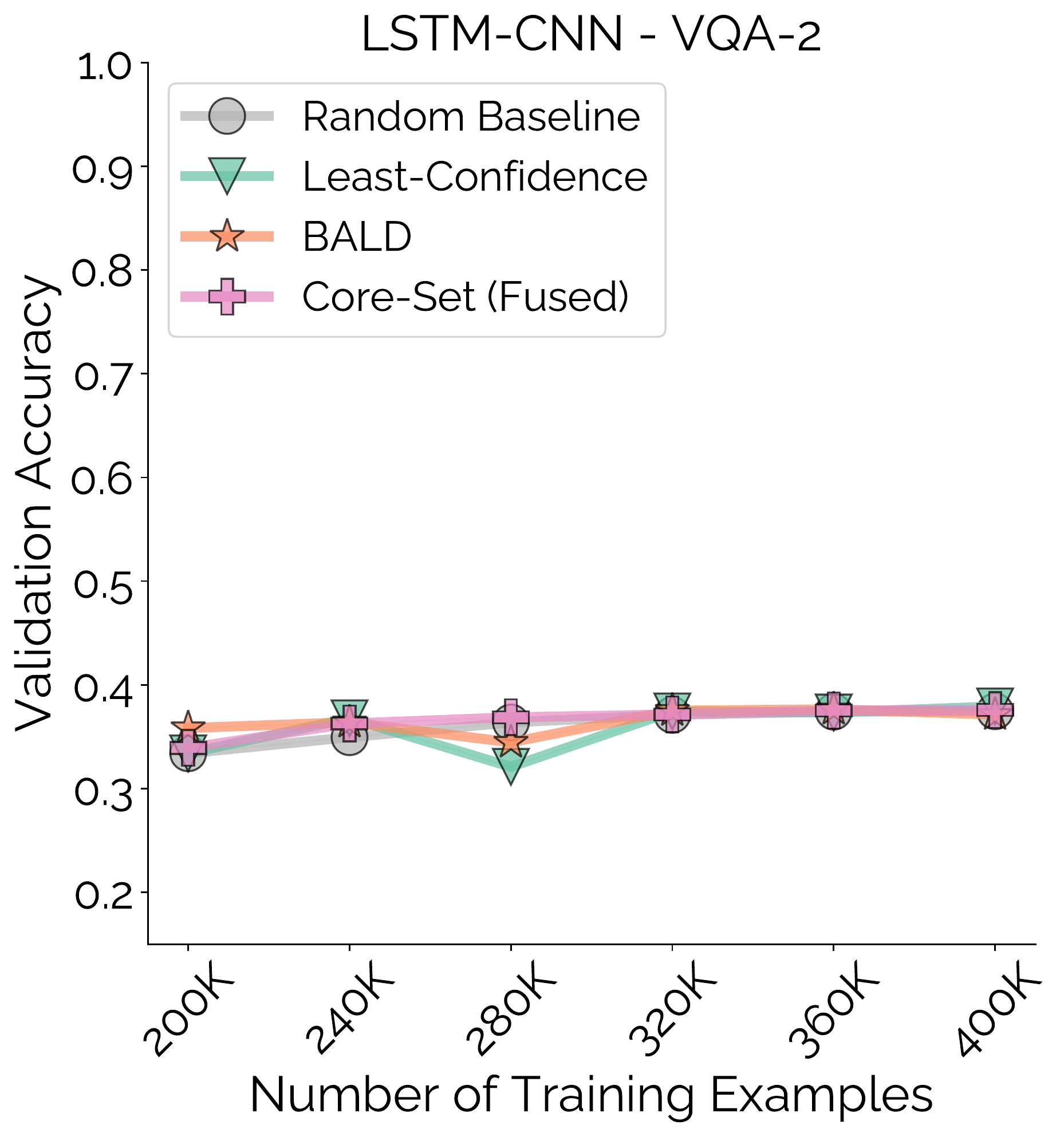}
    \end{subfigure}
    \hfill
    \begin{subfigure}[b]{0.31\textwidth}
        \centering
        \includegraphics[width=\textwidth, height=0.2\textheight]{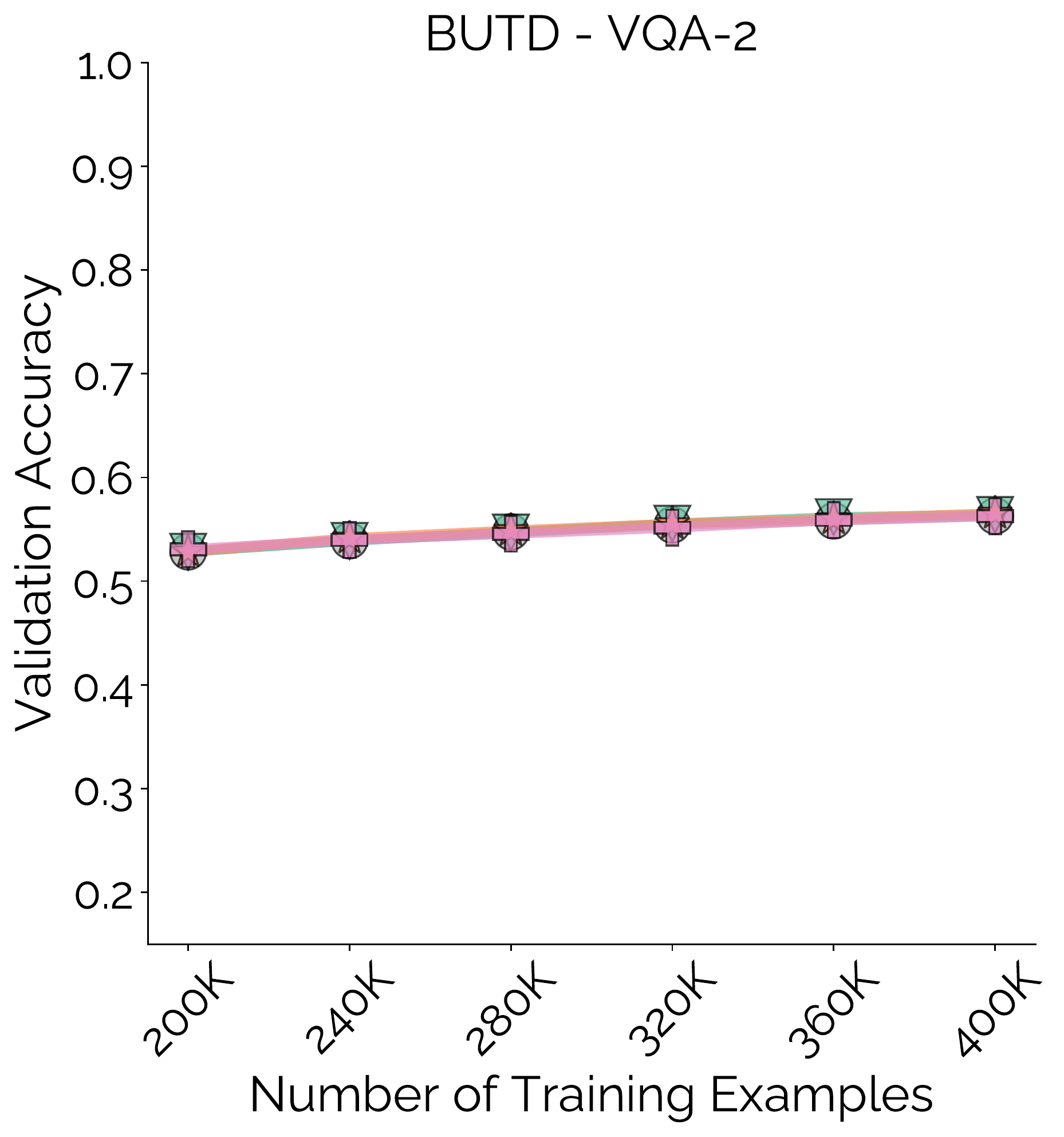}
    \end{subfigure}
    \hfill
    \begin{subfigure}[b]{0.31\textwidth}
        \centering
        \includegraphics[width=\textwidth, height=0.2\textheight]{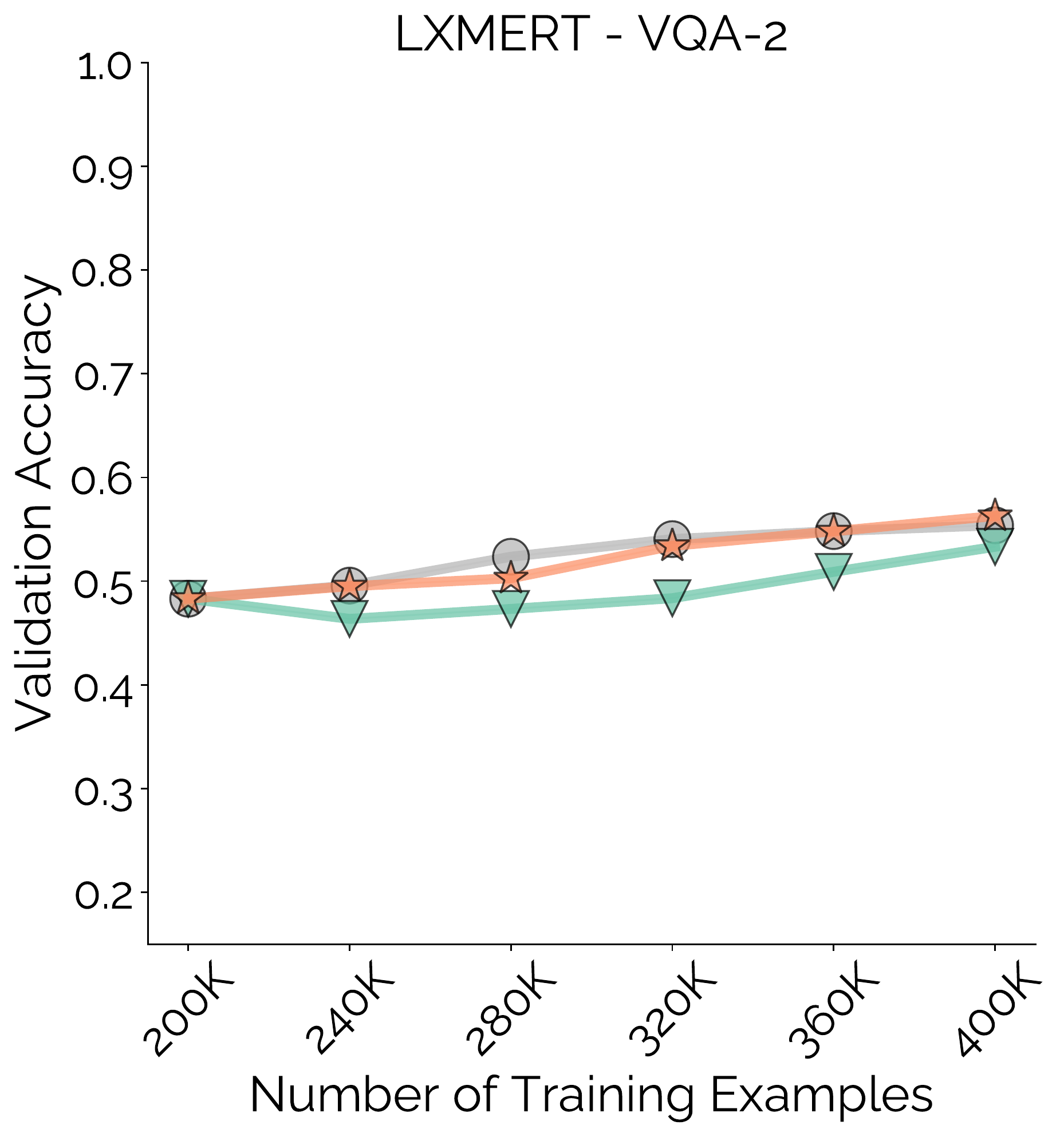}
    \end{subfigure}
    \vspace*{-5pt}
    \caption{Results on VQA-2 using 50\% of the dataset as a seed set. While methods are \textit{relatively} better when using a larger seed set---confirming results from \citep{lin2017active}---no methods outperform random.}    
    \label{fig:vqa2-p50}
\end{figure*}

\begin{figure*}
    \centering
    \begin{subfigure}[b]{0.31\textwidth}
        \centering
        \includegraphics[width=\textwidth, height=0.2\textheight]{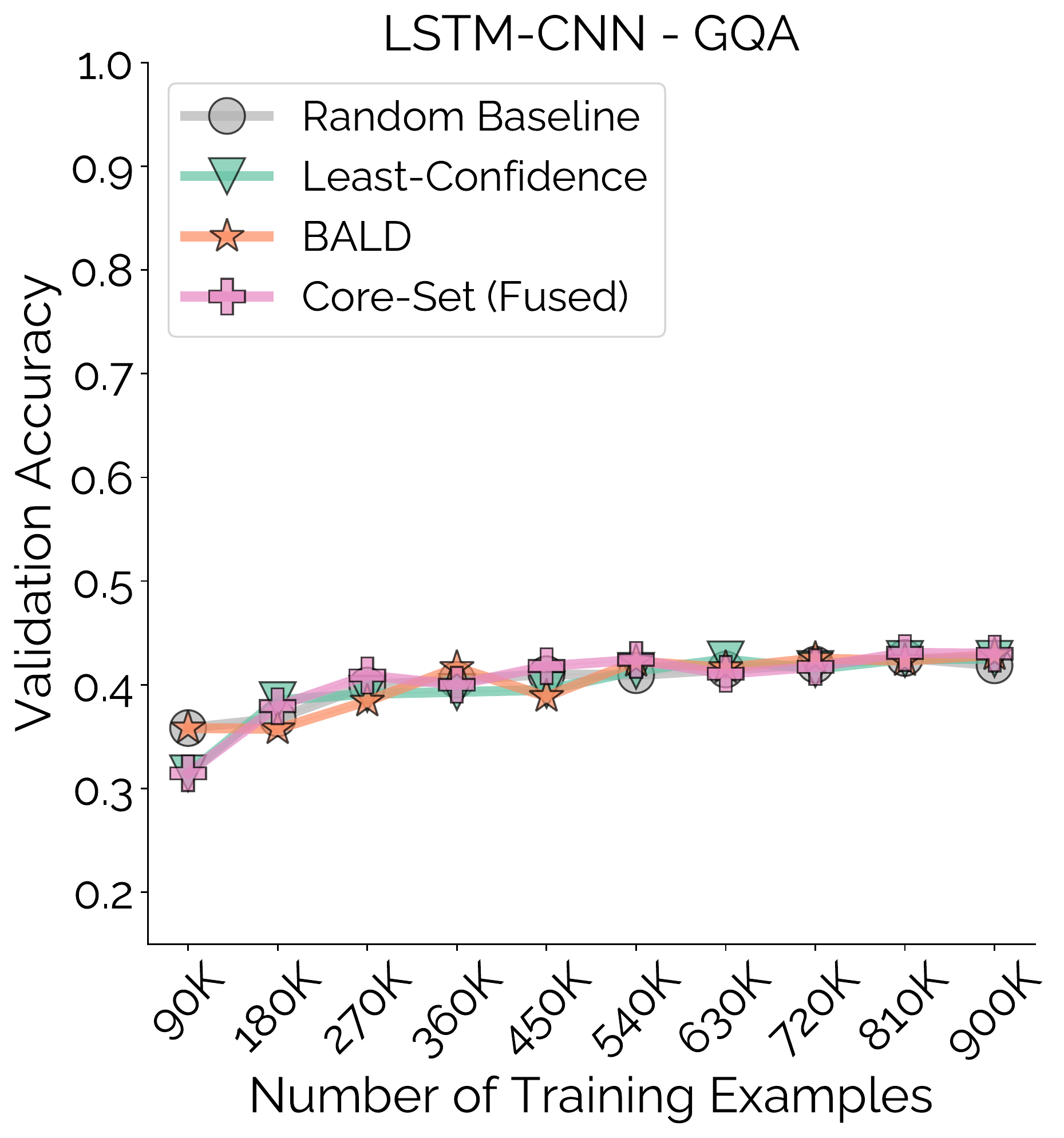}
    \end{subfigure}
    \hfill
    \begin{subfigure}[b]{0.31\textwidth}
        \centering
        \includegraphics[width=\textwidth, height=0.2\textheight]{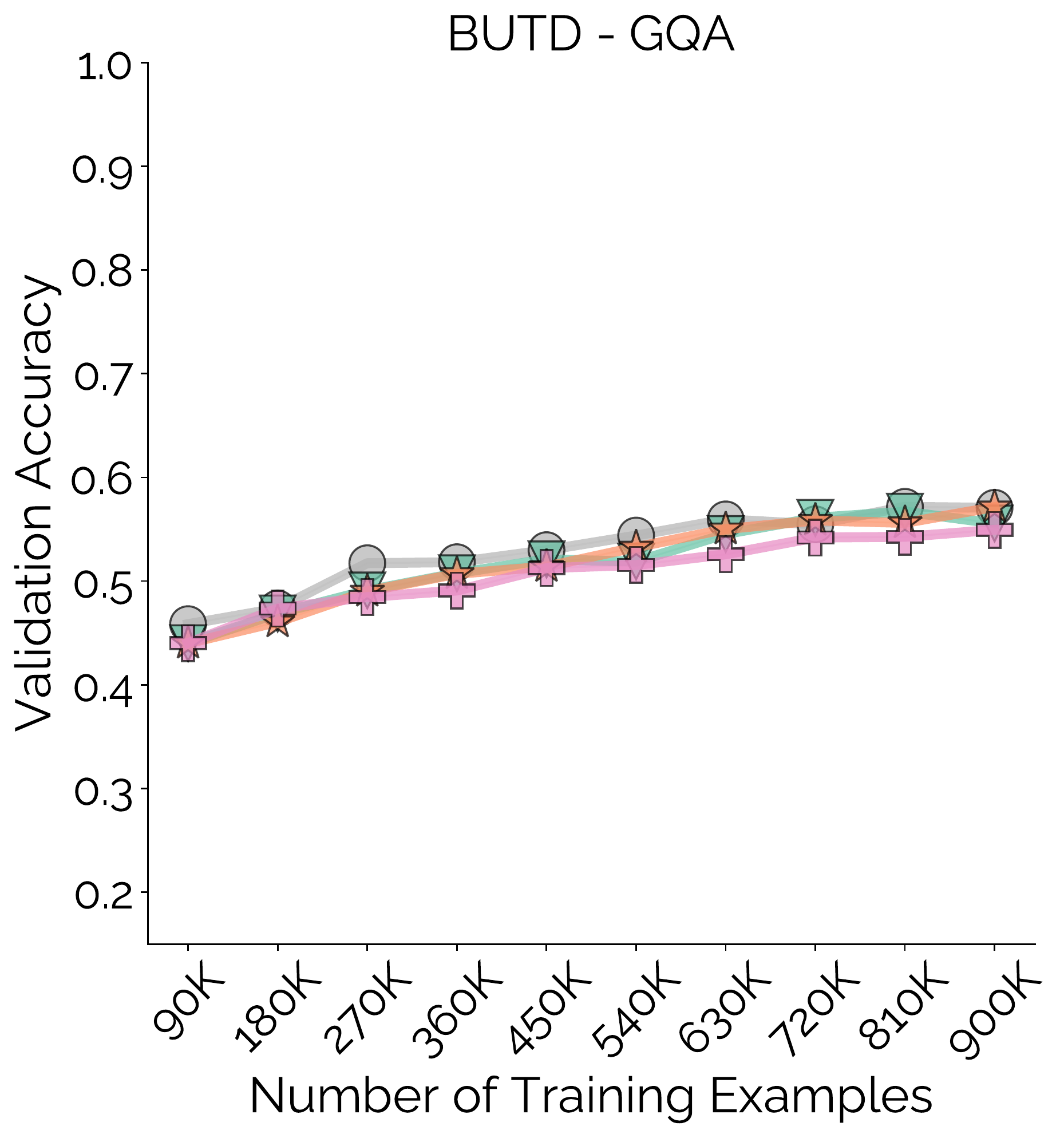}
    \end{subfigure}
    \hfill
    \begin{subfigure}[b]{0.31\textwidth}
        \centering
        \includegraphics[width=\textwidth, height=0.2\textheight]{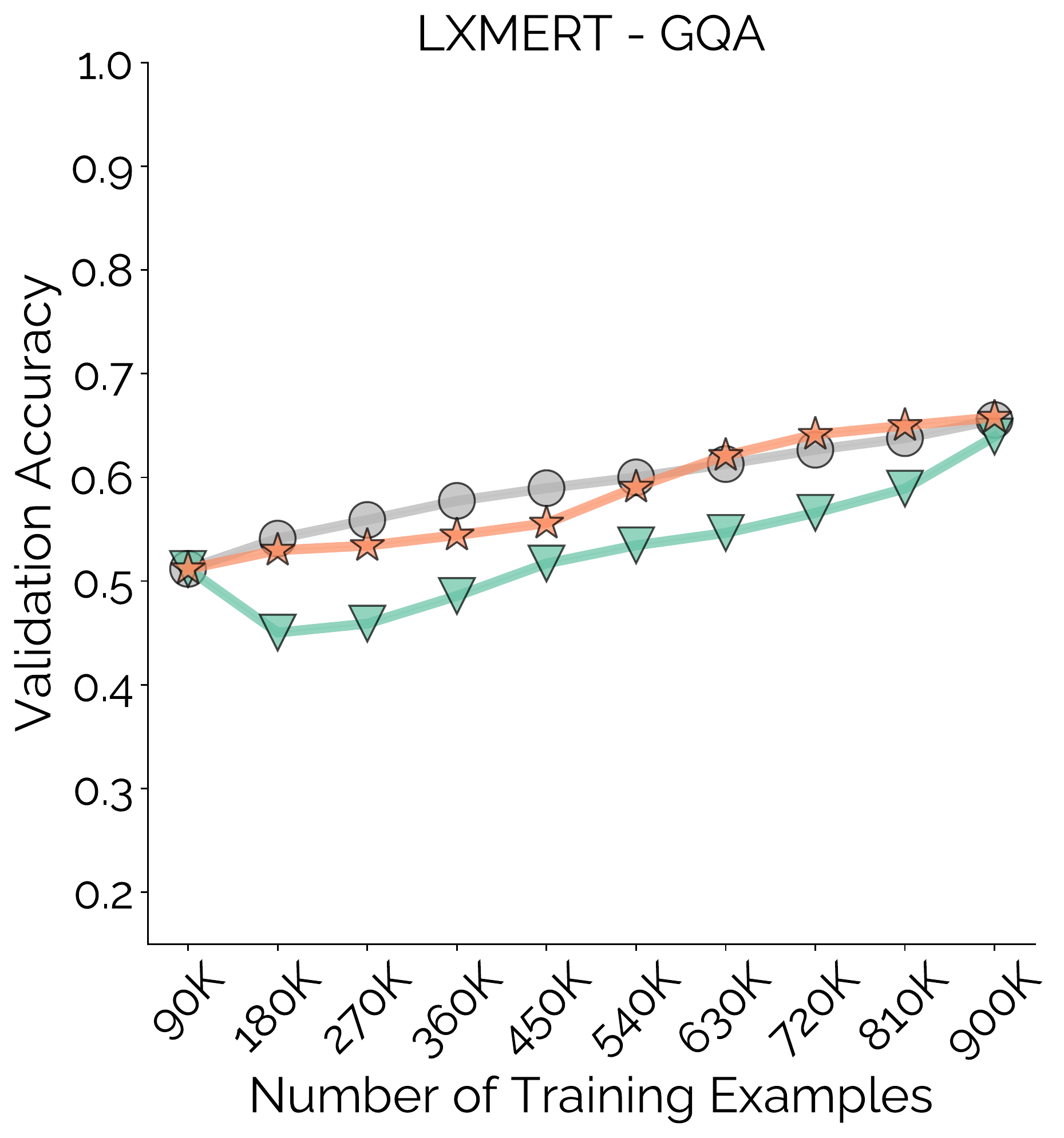}
    \end{subfigure}
    \vspace*{-5pt}
    \caption{Results on \textsc{GQA} using 10\% of the dataset for the seed set. Even with different question structures, the above trends hold, with strategies performing worse than or equivalent to random.}
    \label{fig:gqa-p10}
\end{figure*}

\subsection{Acquisition Functions} 
\label{sec:acquisition-functions}
Several active learning methods have been developed to account for different aspects of the machine learning training pipeline: while some acquire examples with high aleotoric uncertainty \citep{settles2009active} (having to do with the natural uncertainty in the data) or epistemic uncertainty \citep{gal2017dbal} (having to do with the uncertainty in the modeling/learning process), others attempt to acquire examples that reflect the distribution of data in the pool \citep{sener2018active}. We sample a diverse set of these methods:

\paragraph{Random Sampling} serves as our baseline passive approach for acquiring examples.

\paragraph{Least Confidence} acquires examples with lowest model prediction probability \citep{settles2009active}.

\paragraph{Entropy} acquires examples with the highest entropy in the model's output \citep{settles2009active}.

\paragraph{MC-Dropout Entropy} (Monte-Carlo Dropout with Entropy acquisition) acquires examples with high entropy in the model's output averaged over multiple passes through a neural network with different dropout masks \citep{gal2016dropout}. This process is a consequence of a theoretical casting of dropout as approximate Bayesian inference in deep Gaussian processes.

\paragraph{BALD} (Bayesian Active Learning by Disagreement) builds upon Monte-Carlo Dropout by proposing a decision theoretic objective; it acquires examples that maximise the decrease in expected posterior entropy \citep{houlsby2011bayesian,gal2017dbal,siddhant2018deep} -- capturing ``disagreement'' across different dropout masks.

\paragraph{Core-Set Selection} samples examples that capture the diversity of the data pool \citep{sener2018active,coleman2020selection}. It acquires examples to minimize the distance between an example in the unlabeled pool to its closest labeled example. Since Core-Set selection operates over a representation space (and not an output distribution, like prior strategies) and VQA models operate over two modalities, we employ three Core-Set variants: \textbf{Core-Set (Language)} and \textbf{Core-Set (Vision)} operate over their respective representation spaces while \textbf{Core-Set (Fused)} operates over the ``fused'' vision and language representation space.

\begin{figure*}[t]
    \centering
    \begin{subfigure}{0.33\textwidth}
        \centering
        \includegraphics[width=\textwidth, height=0.18\textheight]{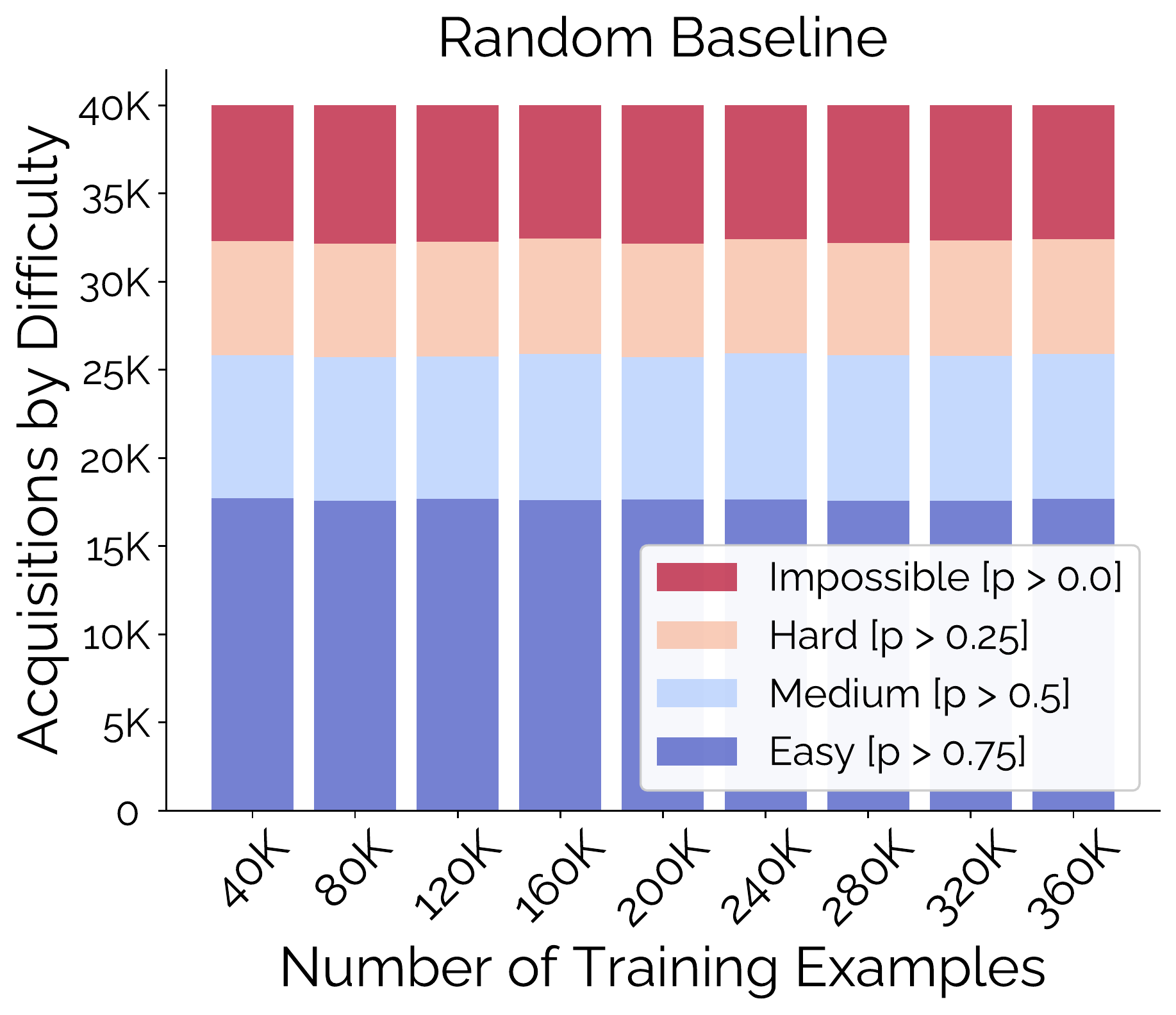}
    \end{subfigure}
    \hspace*{-5pt}
    \hfill
    \begin{subfigure}{0.33\textwidth}
        \centering
        \includegraphics[width=\textwidth, height=0.18\textheight]{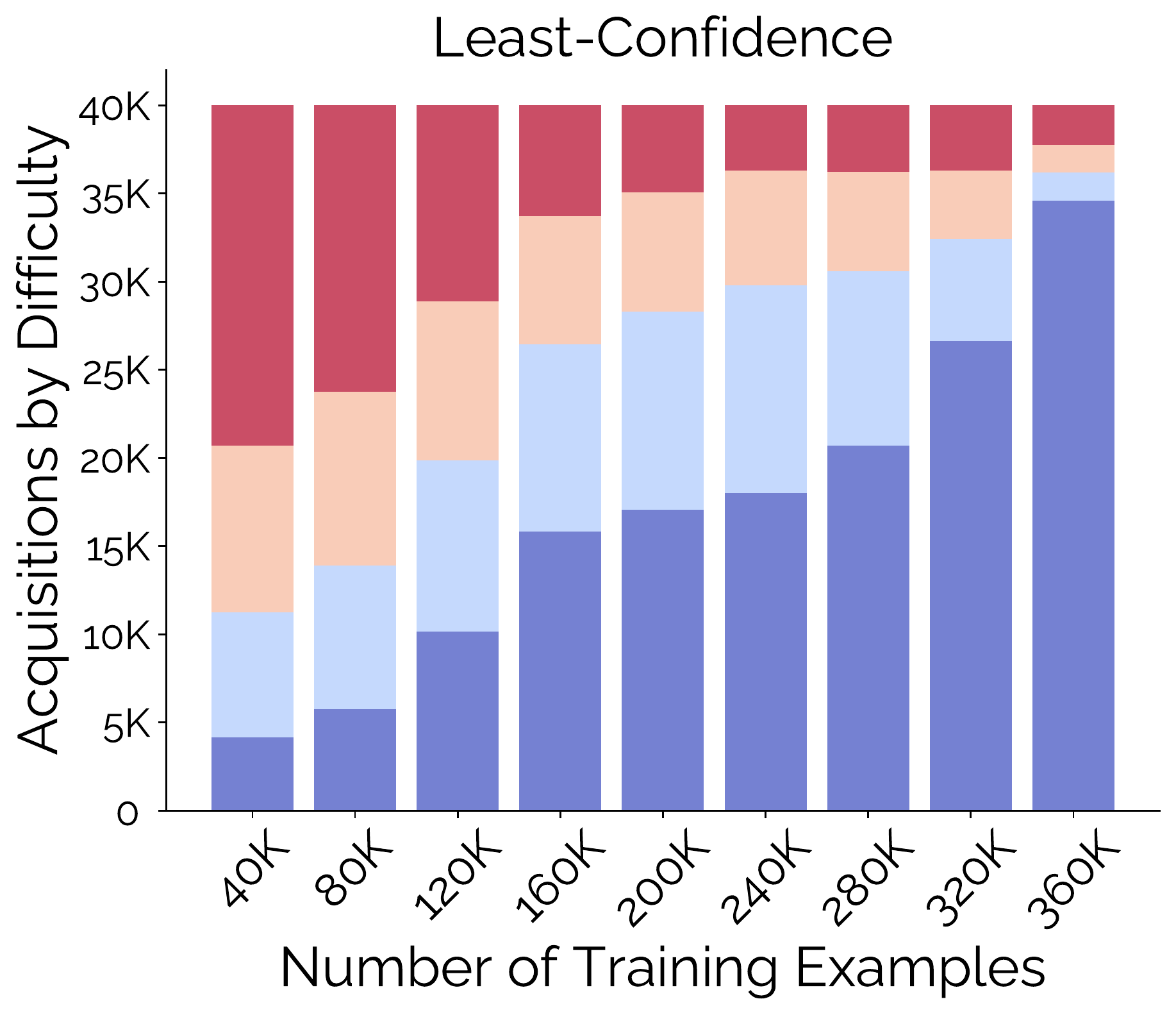}
    \end{subfigure}
    \hspace*{-5pt}
    \hfill
    \begin{subfigure}{0.33\textwidth}
        \centering
        \includegraphics[width=\textwidth, height=0.18\textheight]{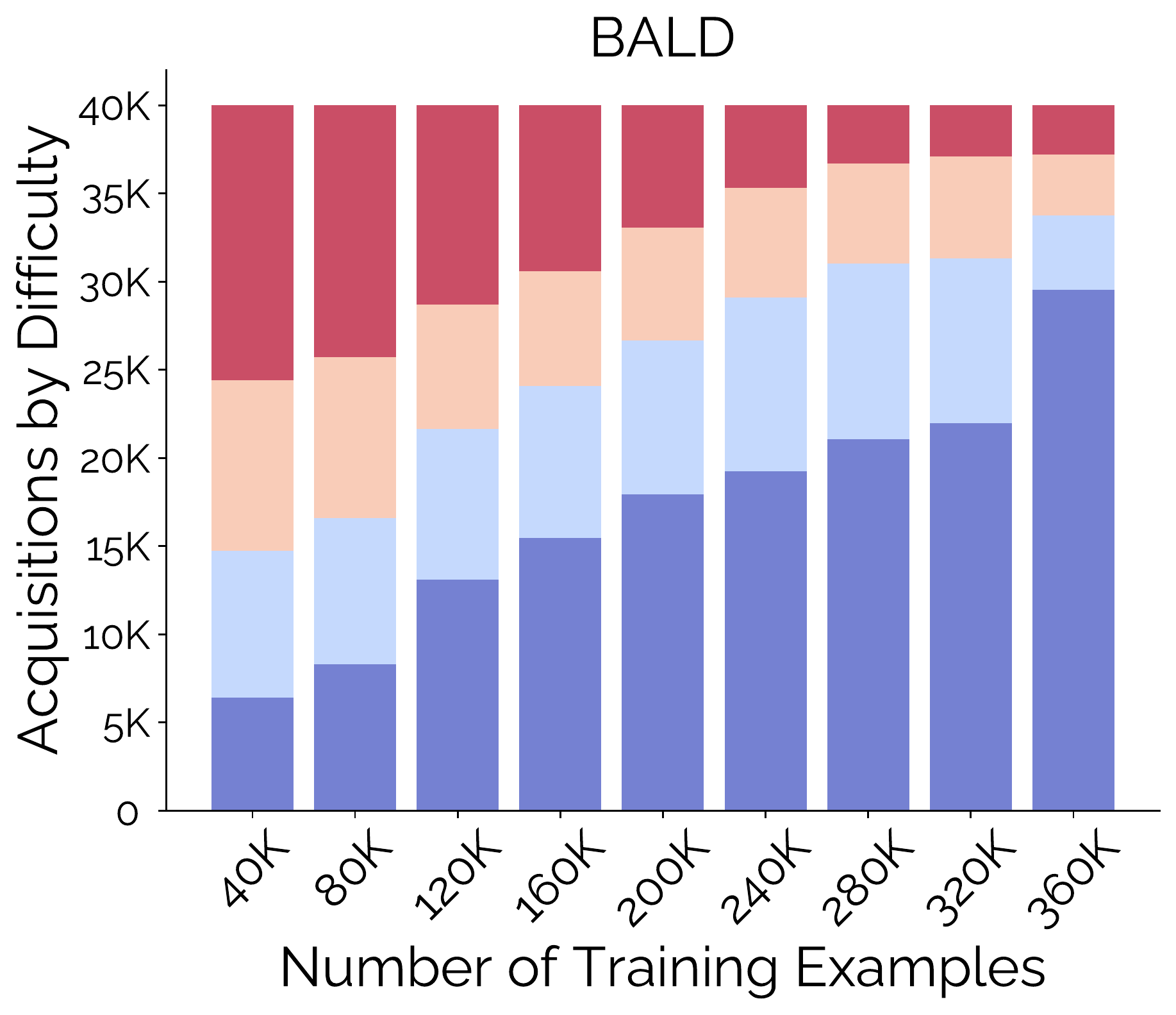}
    \end{subfigure}
    \caption{We visualize the difference in acquisition preferences between random and active learning acquisitions (least confidence and BALD) across multiple iterations. Active learning methods prefer to sample impossible examples which models are unable to learn, hurting sample efficiency relative to the random baseline.}
    \label{fig:acquisitions}
\end{figure*}

\section{Experimental Results}
\label{sec:experiments}
We evaluate the 8 active learning strategies across the 5 models described in the previous section. Figures~\ref{fig:vqa-sports-p10}--\ref{fig:gqa-p10} show a representative sample of active learning results across datasets. Due to space constraints, we only visualize 4 active learning strategies -- Least-Confidence, BALD, CoreSet-Fused, and the Random Baseline -- using 3 models (LSTM-CNN, BUTD, LXMERT).\footnote{For LXMERT, running Core-Set selection is prohibitive, so we omit these results; please see Appendix B for more details.} Results and trends are consistent across the different acquisition functions, models and seed set sizes (see the appendix for results with other models, acquisition functions, and seed set sizes). We now go on to provide descriptions of the datasets we evaluate against, and the corresponding results.

\subsection{Simplified VQA Datasets}
One complexity of VQA is the size of the output space and the number of examples present \citep{agrawal2015vqa1,goyal2017making}; VQA-2 has 400k training examples, and in excess of 3k possible answers (see Table \ref{tab:datasets}). However, prior work in active learning focuses on smaller datasets like the 10-class MNIST dataset \citep{gal2017dbal}, binary classification \citep{siddhant2018deep}, or small-cardinality ($\leq$ 20 classes) text categorization \citep{lowell2019practical}. To ensure our results and conclusions are not due to the size of the output space, we build two meaningful, but narrow-domain VQA datasets from subsets of VQA-2. These simplified datasets reduce the complexity of the underlying learning problem and provide a fair comparison to existing active learning literature.

\paragraph{VQA-Sports.} We generate VQA-Sports by compiling a list of 20 popular sports (e.g.,~soccer, football, tennis, etc.) in VQA-2, and restricting the set of questions to those with answers in this list. We picked the sports categories by ranking the GloVe vector similarity between the word ``sports'' to answers in VQA-2, and selected the 20 most commonly occurring answers. 

\paragraph{VQA-Food.} We generate the VQA-Food dataset similarly, compiling a list of the 20 commonly occurring food categories by GloVe vector similarity to the word ``food.''

\paragraph{Results.} Figure \ref{fig:vqa-sports-p10} presents results for VQA-Sports, with an initial seed set restricted to 10\% of the total pool (500 examples). The appendix reports similar results on VQA-Food. For LSTM-CNN, \textit{Least-Confidence} appears to be slightly more sample efficient, while all other strategies perform on par with or worse than random. For BUTD, all methods are on par with random; for LXMERT, they perform worse than random. Generally on VQA-Sports, active learning performance varies, but fails to outperform random acquisition.

\subsection{VQA-2} 
VQA-2 is the canonical dataset for evaluating VQA models \citep{goyal2017making}. In keeping with prior work \citep{anderson2018butd, tan2019lxmert}, we filter the training set to only include answers that appear at least 9 times, resulting in 3130 unique answers. Unlike traditional VQA-2 evaluation, which treats the task as a \textit{multi-label} binary classification problem, we follow prior active learning work on VQA~\cite{lin2017active}, which formulates it as a \textit{multi-class} classification problem, enabling the use of acquisition functions such as uncertainty sampling and BALD.

\paragraph{Results.} Figures~\ref{fig:vqa2-p10} and \ref{fig:vqa2-p50} show results on VQA-2 with different seed set sizes -- 10\% (40k examples) and 50\% (200k examples). Active learning performs relatively better with larger seed sets but still underperforms random. Surprisingly, when initialized with 50\% of the pool as the seed set, the gain in validation accuracy after acquiring the entire pool of examples (400k examples total) is only 2\%. This is an indication that the lack of sample efficiency might be a result of the underlying data, a problem we explore in the next section.

\begin{figure*}[t!]
    \centering
    \includegraphics[width=.95\linewidth]{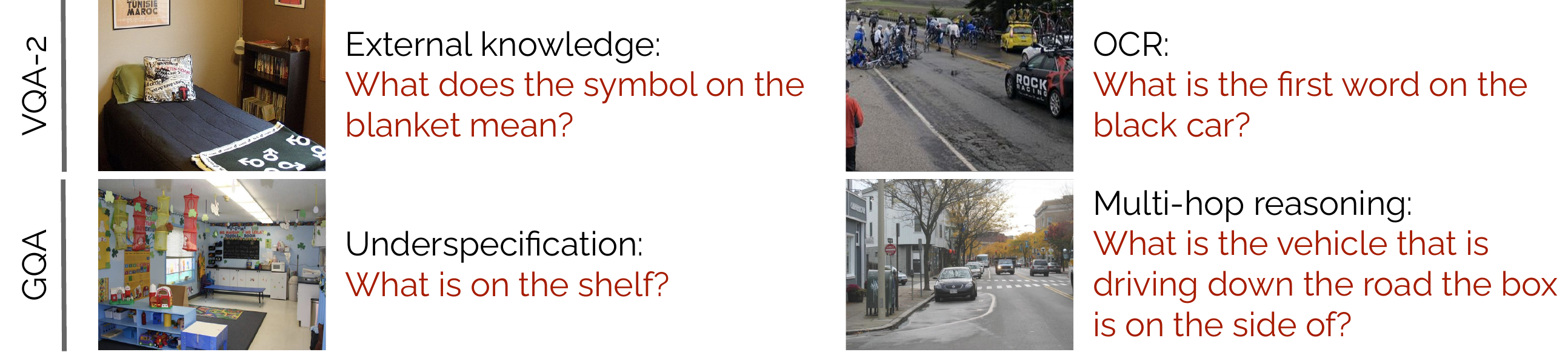}
    \caption{Example groups of collective outliers in the VQA-2 and GQA datasets.}
    \label{fig:outlier_examples}
    \vspace*{-10pt}
\end{figure*}

\subsection{GQA}
GQA was introduced as a means for evaluating compositional reasoning \citep{hudson2019gqa}. Unlike VQA's natural human-written questions, GQA contains synthetic questions of the form ``what is inside the bottle the glasses are to the right of?''. We use the standard GQA training set of 943k questions, 900k of which we use for the active learning pool.

\paragraph{Results.} Figure \ref{fig:gqa-p10} shows results on GQA using a seed set of 10\% of the full pool (90k examples). Despite its notable differences in question structure to VQA-2, active learning still performs on par with or slightly worse than random.

\section{Analysis via Dataset Maps}
\label{sec:cartography}
The previous section shows that active learning fails to improve over random acquisition on VQA across models and datasets. A simple question remains -- \textit{why}? One hypothesis is that sample inefficiency stems from the data itself: there is only a 2\% gain in validation accuracy when training on half versus the whole dataset. Working from this, we characterize the underlying datasets using Dataset Maps \citep{swayamdipta2020dataset} and discover that active learning methods prefer sampling ``hard-to-learn'' examples, leading to poor performance.

\paragraph{Mapping VQA Datasets.}
A Dataset Map \citep{swayamdipta2020dataset} is a model-specific graph for profiling the learnability of individual training examples. Dataset Maps present holistic pictures of classification datasets relative to the training dynamics of a given model; as a model trains for multiple epochs and sees the same examples repeatedly, the mapping process logs statistics about the confidence assigned to individual predictions. Maps then visualize these statistics against two axes: the y-axis plots the average model confidence assigned to the correct answer over training epochs, while the x-axis plots the spread, or variability, of these values. This introduces a 2D representation of a dataset (viewed through its relationship with individual model) where examples are placed on the map by coarse statistics describing their ``learnability``. We show the Dataset Map for BUTD trained on VQA-2 in Figure~\ref{fig:pull_figure}. For our work, we build this map post-hoc, training on the entire pool as a means for analyzing what active learning is doing -- treating it as a diagnostic tool for identifying the root cause why active learning seems to fail for VQA.

In an ideal setting, the majority of examples in the training set should lie in the upper half of the graph -- i.e., the mean confidence assigned to the correct answer should be relatively high. Examples towards the upper-left side represent the ``easy-to-learn'' examples, as the variability in the confidence assigned by the model over time is fairly low.

A curious feature of VQA-2 and other VQA datasets is the presence of the 25-30\% of examples in the bottom-left of the map (shown in red in Figure~\ref{fig:pull_figure}) -- examples that have low confidence and variability. In other words, models are unable to learn a large proportion of training examples. While prior work attributes examples in this quadrant to ``labeling errors'' \citep{swayamdipta2020dataset}, labeling errors in VQA are sparse, and cannot account for the density of such examples in these maps.

\begin{figure*}[t]
    \centering
    \begin{subfigure}{0.31\textwidth}
        \centering
        \includegraphics[width=\textwidth, height=0.2\textheight]{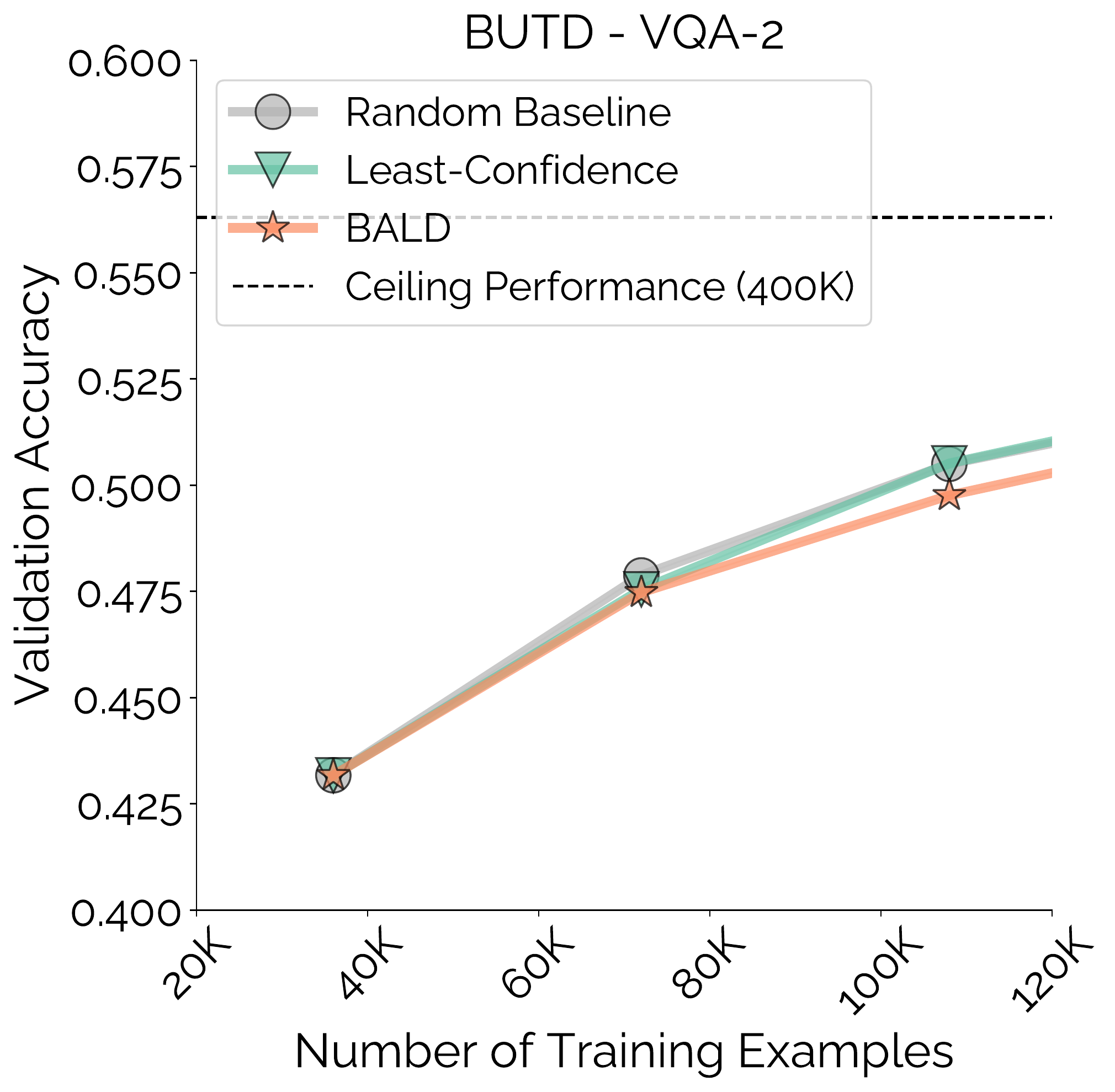}
        \caption{10\% of Dataset Removed}
    \end{subfigure}
    \hfill
    \begin{subfigure}{0.31\textwidth}
        \centering
        \includegraphics[width=\textwidth, height=0.2\textheight]{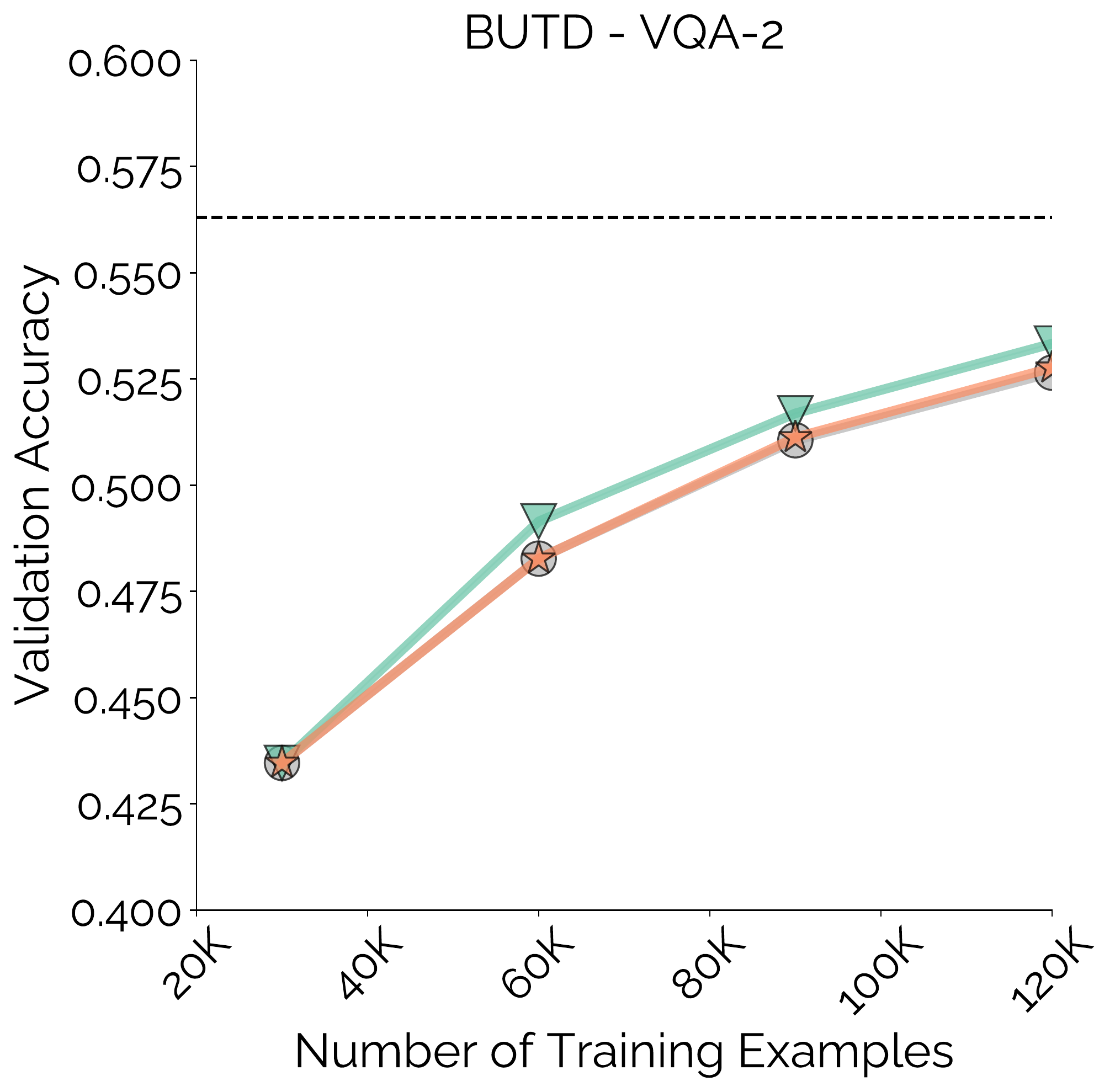}
        \caption{25\% of Dataset Removed}
    \end{subfigure}
    \hfill
    \begin{subfigure}{0.31\textwidth}
        \centering
        \includegraphics[width=\textwidth, height=0.2\textheight]{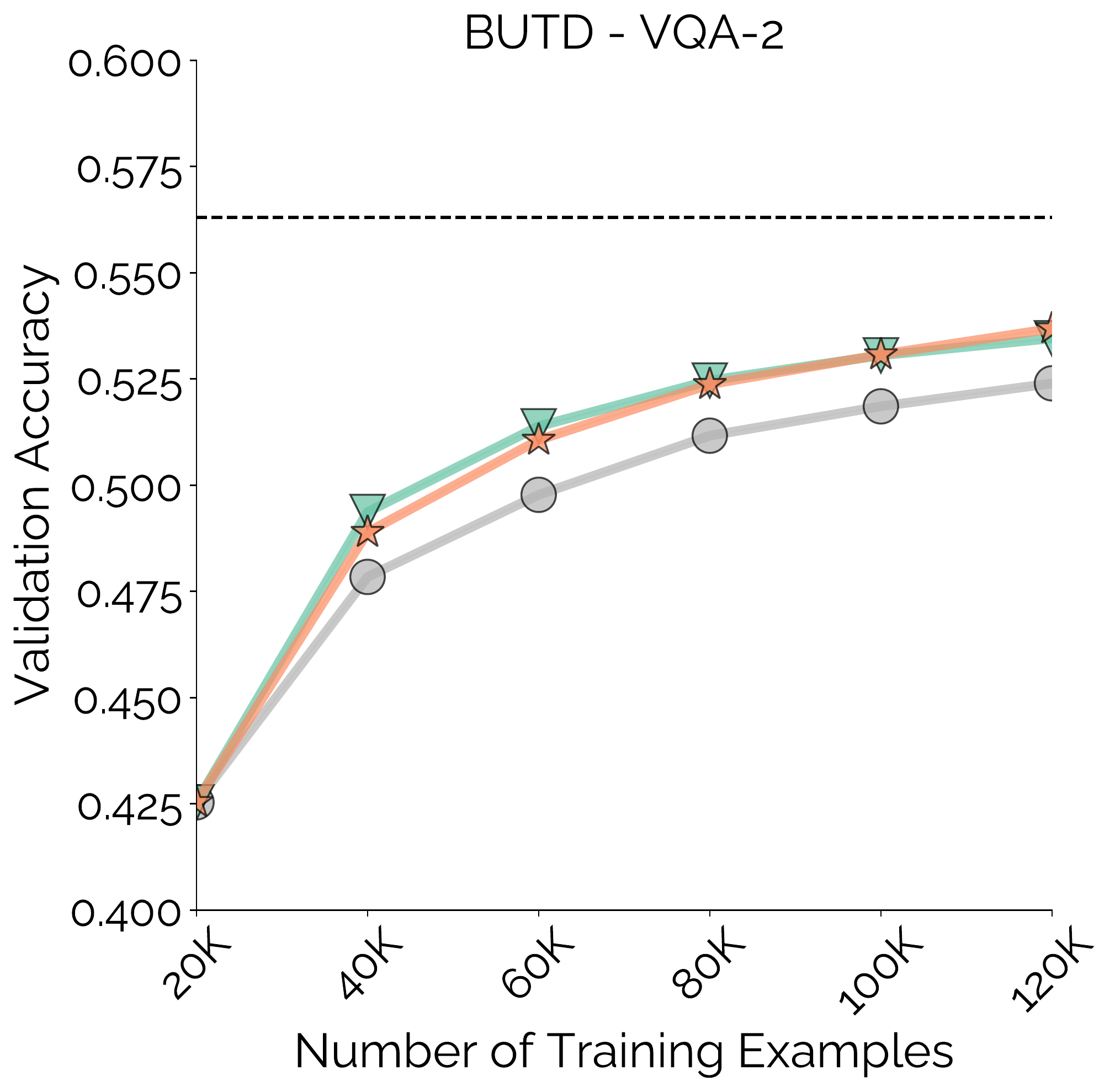}
        \caption{50\% of Dataset Removed}
    \end{subfigure}
    \caption{Using Dataset Maps, we remove hard-to-learn examples, which we identify as collective outliers. With the outliers removed, active learning methods demonstrate up to 2--3x sample efficiency versus random sampling.}
    \label{fig:outlier-ablations}
\end{figure*}

\paragraph{Interpreting Acquisitions.}
We profile the acquisitions made by each active learning method, contextualizing the acquired examples via their placement on the associated Dataset Map. We segregate training examples into four buckets using the map's y-axis: easy ($\ge 0.75$), medium ($\ge 0.50$), hard ($\ge 0.25$), and impossible ($\ge 0.00$). Ideally, active learning should be robust to ``hard-to-learn'' examples, focusing instead on learnable, high uncertainty examples towards the upper-right portion of the Dataset Map. Instead, we find that active learning methods acquire a large proportion of impossible examples early on and concentrate on the easier examples only after the impossible examples dwindle (see Figure~\ref{fig:acquisitions}). In contrast, the random baseline acquires examples proportional to each bucket's density in the underlying map; acquiring easier examples earlier and performing on par with or better than all others.

\section{Collective Outliers}
\label{sec:outliers}
This leaves two questions: 1) can we characterize these ``hard'' examples, and 2) are these examples responsible for the ineffectiveness of active learning on VQA? We first identify hard-to-learn examples as collective outliers and explain why active learning methods prefer to acquire them. Next, we perform ablation experiments, removing these outliers from the active learning pool iteratively, and demonstrate a corresponding boost in sample efficiency relative to random acquisition. 

\paragraph{Hard Examples are Collective Outliers.}
Collective outliers are groups of examples that deviate from the rest of the examples but cluster together \citep{han2000datamining} -- they often present as fundamental subproblems of a broader task. For instance (Figure~\ref{fig:outlier_examples}), in VQA-2, we identify clusters of hard-to-learn examples that require optical character recognition (OCR) for reasoning about text (e.g., ``What is the first word on the black car?''); another cluster requires external knowledge to answer (``What is the symbol on the hood often associated with?''). In GQA, we identify different clusters of collective outliers; one cluster stems from innate underspecification (e.g., ``what is on the shelf?'' with multiple objects present on the shelf); another cluster requires multiple reasoning hops difficult for current models (e.g., ``What is the vehicle that is driving down the road the box is on the side of?'').

We sample 100 random ``hard-to-learn'' examples from both VQA-2 and GQA and find that 100\% of the examples belong to one of the two aforementioned collectives. Since hard-to-learn examples constitute 25--30\% of the data pool, active learning methods cannot avoid them. Uncertainty-based methods (e.g., Least-Confidence, Entropy, Monte-Carlo Dropout) identify them as valid acquisition targets because models lack the capacity to correctly answer these examples, assigning low confidence and high uncertainty. Disagreement-based methods (e.g., BALD) are similar; model confidence is generally low but high variance (lower middle/lower right of the Dataset Maps). Finally, diversity methods (e.g., Core-Set selection) identify these examples as different enough from the existing pool to warrant acquisition, but fail to learn meaningful representations, fueling a vicious cycle wherein they continue to pick these examples. 

\paragraph{Ablating Outliers.}
To verify that collective outliers are responsible for the degradation of active learning performance, we re-run our experiments using active learning pools with varying numbers of outliers removed. To remove these outliers, we sort and remove all examples in the data pool using the product of their model confidence and prediction variability (x and y-axis values of the Dataset Maps). We systematically remove examples with a low product value and observe how active learning performance changes (see Figure~\ref{fig:outlier-ablations}).

We observe a 2--3x improvement in sample efficiency when removing 50\% of the entire data pool, consisting mainly of collective outliers (Figure \ref{fig:outlier-ablations}c). This improvement decreases if we only remove 25\% of the full pool (Figure \ref{fig:outlier-ablations}b), and further degrades if we remove only 10\% (Figure \ref{fig:outlier-ablations}a). This ablation demonstrates that active learning methods are more sample efficient than the random baseline when collective outliers are absent from the unlabelled pool.
\section{Discussion and Future Work}
\label{sec:discussion}
This paper asks a simple question -- why does the modern neural active learning toolkit fail when applied to complex, open ended tasks? While we focus on VQA, collective outliers are abundant in tasks such as natural language inference \citep{bowman2015large,williams2018broad} and open-domain question answering \citep{kwiatkowski2019natural}, amongst others. More insidious is their nature; collective outliers can take multiple forms, requiring external domain knowledge or ``commonsense'' reasoning, containing underspecification, or requiring capabilities beyond the scope of a given model (e.g., requiring OCR ability). While we perform ablations in this work removing collective outliers, demonstrating that active learning fails as collective outliers take up larger portions of the dataset, this is only an analytical tool; these outliers are, and will continue to be, pervasive in open-ended datasets -- and as such, we will need to develop better tools for learning (and performing active learning) in their presence.

\paragraph{Selective Classification.} One potential direction for future work is to develop systems that abstain when they encounter collective outliers. Historical artificial intelligence systems, such as SHRDLU \citep{winograd1972language} and QUALM \citep{lehnert1977process}, were designed to flag input sequences that they were not designed to parse. Ideas from those methods can and should be resurrected using modern techniques; for example, recent work suggests that a simple classifier can be trained to identify out-of-domain data inputs, provided a seed out-of-domain dataset \citep{kamath2020squads}. Active learning methods can be augmented with a similar classifier, which re-calibrates active learning uncertainty scores with this classifier's predictions. Other work learns to identify novel utterances by learning to intelligently set thresholds in representation space \citep{karamcheti2020decomposition}, a powerful idea especially if combined with other representation-centric active learning methods like Core-Set Sampling \citep{sener2018active}.

\paragraph{Active Learning with Global Reasoning.} Another direction for future work to explore is to leverage Dataset Maps to perform more global, holistic reasoning over datasets, to intelligently identify promising examples -- in a sense, baking part of the analysis done in this work directly into the active learning algorithms. A possible instantiation of this idea would be in training a discriminator to differentiate between ``learnable'' examples (upper half of each Dataset Map) from the ``unlearnable'', collective outliers with low confidence and low variability. Between each active learning acquisition iteration, one can generate an updated Dataset Map, thereby reflecting what models are learning as they obtain new labeled examples.

Machine learning systems deployed in real-world settings will inevitably encounter open-world datasets, ones that contain a mixture of learnable and unlearnable inputs. Our work provides a framework to study when models encounter such inputs. Overall, we hope that our experiments serve as a catalyst for future work on evaluating active learning methods with inputs drawn from open-world datasets.

\section*{Reproducibility}
\label{sec:reproducibility}
All code for data preprocessing, model implementation, and active learning algorithms is made available at \url{https://github.com/siddk/vqa-outliers}. Additionally, this repository also contains the full set of results and dataset maps as well. 

\medskip

\noindent 
The authors are fully committed to maintaining this repository, in terms of both functionality and ease of use, and will actively monitor both email and Github Issues should there be problems.

\section*{Acknowledgements}
\label{sec:acknowledgements}
We thank Kaylee Burns, Eric Mitchell, Stephen Mussman, Dorsa Sadigh, and our anonymous ACL reviewers for their useful feedback on earlier versions of this paper. We are also grateful to Hao Tan for providing us with the LXMERT checkpoint trained without access to VQA datasets, as well as for general LXMERT fine-tuning pointers. 

\medskip

\noindent
Siddharth Karamcheti is graciously supported by the Open Philanthropy Project AI Fellowship. Christopher D. Manning is a CIFAR Fellow.

\bibliographystyle{acl_natbib}
\bibliography{all}

\clearpage
\appendix

\section{Overview}

Due to the broad scope of our experiments and analysis, we were unable to fit all our results in the main body of the paper. Furthermore, given the limited length provided by the appendix, we provide only salient implementation details and other representative results here; however, we make all code, models, data, results, active learning implementations available at this link: \url{https://github.com/siddk/vqa-outliers}.

Generally, any combination of $\{$\textit{active learning strategy} $\times$ \textit{model} $\times$ \textit{seed set size} $\times$ \textit{analysis/acquisition plot}$\}$ is present in this paper, and is available in the public code repository.

\section{Implementation Details}

\subsection{Models \& Training}

Where applicable, we implement our models based on publicly available PyTorch implementations. For the LSTM-CNN model, we base our implementation off of this repository: \url{https://github.com/Shivanshu-Gupta/Visual-Question-Answering}, while for the Bottom-Up Top-Down Attention Model, we use this repository: \url{https://github.com/hengyuan-hu/bottom-up-attention-vqa}, keeping default hyperparameters the same.

\paragraph{Logistic Regression.} When implementing Logistic Regression, we base our PyTorch implementation on the broadly used Scikit-Learn (\url{https://scikit-learn.org}) implementation, using the default parameters (including L2 weight decay). We optimize our models via stochastic gradient descent.

\paragraph{LXMERT.} As mentioned in Section 3, the default LXMERT checkpoint and fine-tuning code made publicly available in \citet{tan2019lxmert} (associated code repository: \url{https://github.com/airsplay/lxmert}) is pretrained on data from VQA-2 and GQA, leaking information that could substantially affect our active learning results. To mitigate this, we contacted the authors, who kindly provided us with a checkpoint of the model without VQA pretraining.

However, in addition to this model obtaining different results from those reported in the original work, the provided pretrained checkpoint behaves slightly differently during fine-tuning, requiring different hyperparameters than provided in the original repository. We perform a coarse grid search over hyperparameters, using the LXMERT implementation provided by HuggingFace Transformers \citep{wolf2019transformers}, and find that using an AdamW optimizer rather than the BERT-Adam Optimizer used in the original work \textit{without any special learning rate scheduling} results in the best fine-tuning performance.

\subsection{Acquisition Functions}

We use standard implementations of the 8 active learning strategies described, borrowing from prior implementations \citep{mussmann2018accuracy} and existing code repositories (\url{https://github.com/google/active-learning}). We provide additional details below.

\paragraph{Monte-Carlo Dropout.} For our implementations of the deep Bayesian active learning methods (Monte-Carlo Dropout w/ Entropy, BALD), we follow \citet{gal2016dropout} and estimate a Dropout distribution via test-time dropout, running multiple forward passes through our neural networks, with different, randomly sampled Dropout masks. We use a value of $k=10$ forward passes to form our Dropout distribution.

\paragraph{Amortized Core-Set Selection.} In the original Core-Set selection active learning work introduced by \citet{sener2018active}, it is shown that Core-Set selection for active learning can be reduced to a version of the \textit{k-centers problem}, which can be solved approximately (2-OPT) with a greedy algorithm. However, running this algorithm on high-dimensional representations, across large pools can be prohibitive; Core-Set selection is \textit{batch-aware}, requiring recomputing distances from each ``cluster-center'' (points in the set of acquired examples) to all points in the active learning pool \textit{after each acquisition in a batch}. While we can run this out completely for smaller datasets (and indeed, this is what we do for our small datasets VQA-Sports and VQA-Food), a single acquisition iteration for a large dataset for the full VQA-2 dataset takes approximately 20 GPU-hours on the resources we have available, or up to 9 days for a single Core-Set selection run. For GQA, performing exact Core-Set selection takes at least twice as long.

To still capture the spirit of Core-Set diversity-based selection in our evaluation, we instead introduce an \textit{amortized implementation of Core-Set selection}, which is comprised of two steps. We first downsample the high-dimensional representations (of either the fused language and text, or either unimodal representations) via Principal Component Analysis (PCA) to make the distance computation faster by an order of magnitude. Then, rather than updating distances from examples in our acquired set to points in our pool \textit{after each acquisition $\hat{x}$}, we delay updates, instead only refreshing the distance computation every 2000 acquisitions (roughly 5\% of an acquisition batch for VQA-2). This allows us to report results for Core-Set selection with the three different proposed representations (Fused, Language-Only, Vision-Only) for VQA-2; unfortunately, for GQA and LXMERT (due to the high cost of training), even running this amortized version of Core-Set selection is prohibitive, so we report a subset of results, and omit the rest.

\section{Active Learning Results}

We include further results from our study of active learning applied to VQA, including results on VQA-Food (not included in the main body), active learning results for the two logistic regression models -- Log-Reg (ResNet-101) and Log-Reg (Faster R-CNN), as well as with the 4 acquisition strategies not included in the main body of the paper -- Entropy, Monte-Carlo Dropout w/ Entropy, Core-Set (Language), and Core-Set (Vision).

\subsection{VQA-Food}

\begin{figure*}
    \centering
    \begin{subfigure}[b]{0.31\textwidth}
        \centering
        \includegraphics[width=\textwidth, height=0.2\textheight]{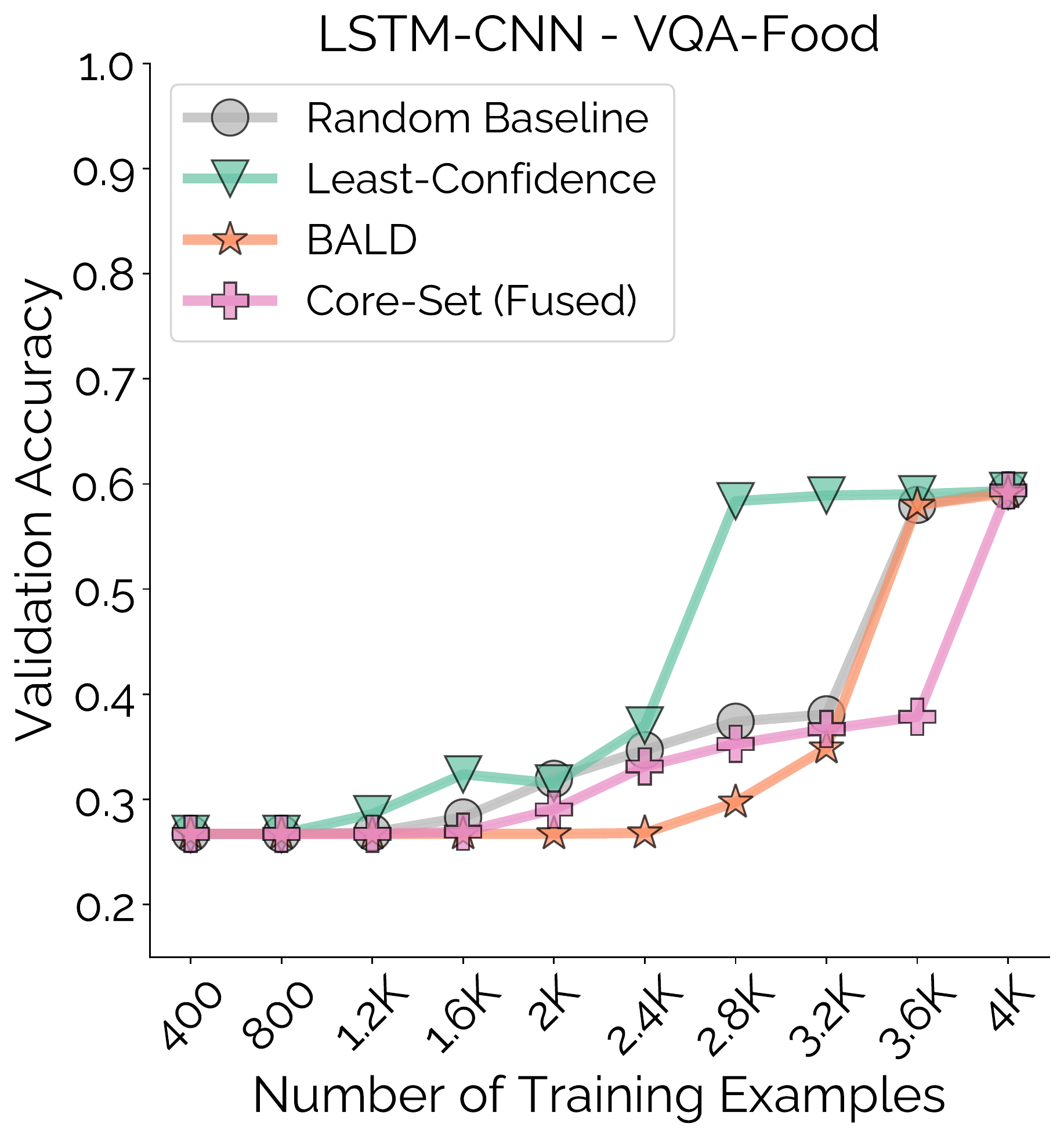}
    \end{subfigure}
    \hfill
    \begin{subfigure}[b]{0.31\textwidth}
        \centering
        \includegraphics[width=\textwidth, height=0.2\textheight]{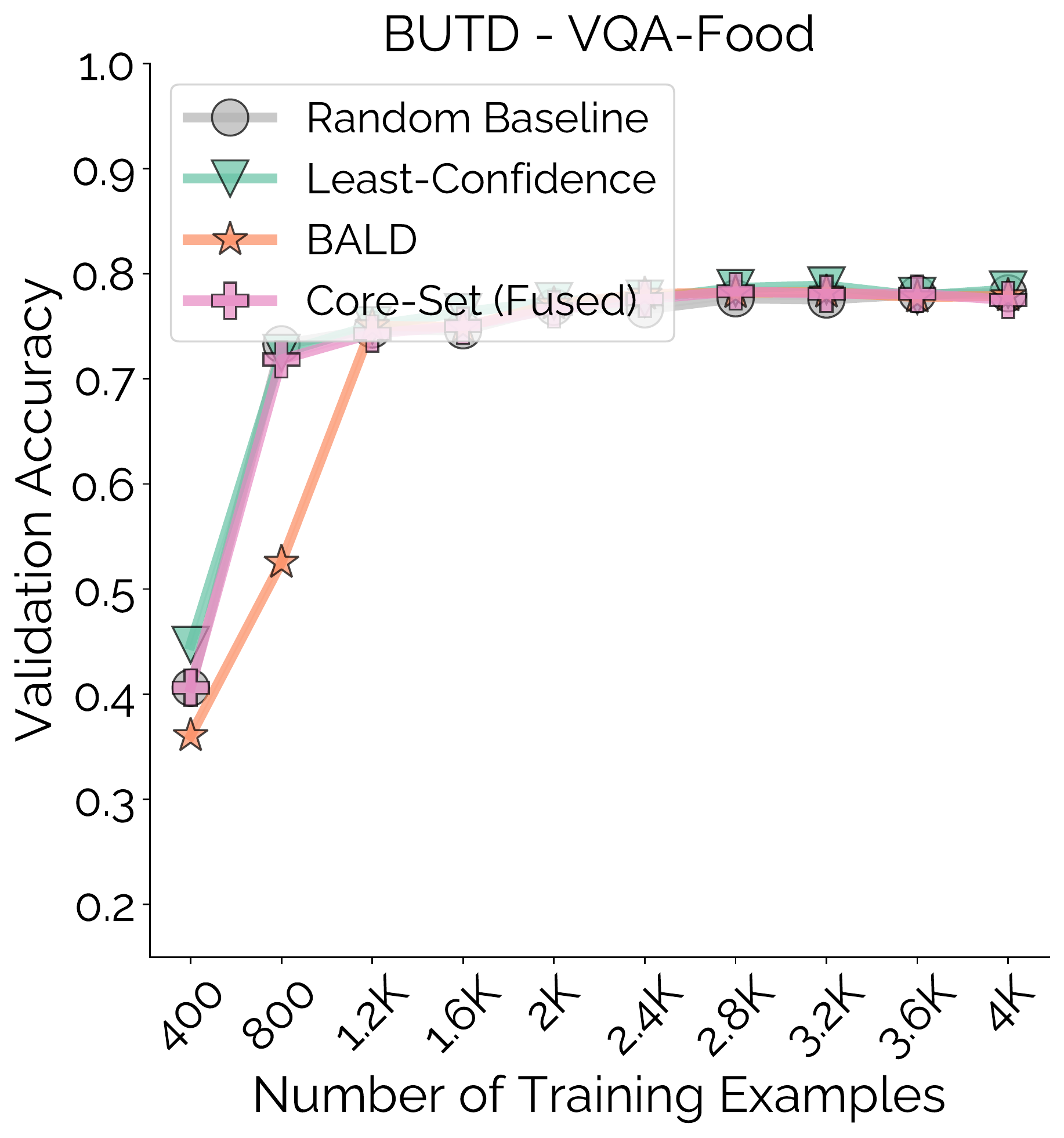}
    \end{subfigure}
    \hfill
    \begin{subfigure}[b]{0.31\textwidth}
        \centering
        \includegraphics[width=\textwidth, height=0.2\textheight]{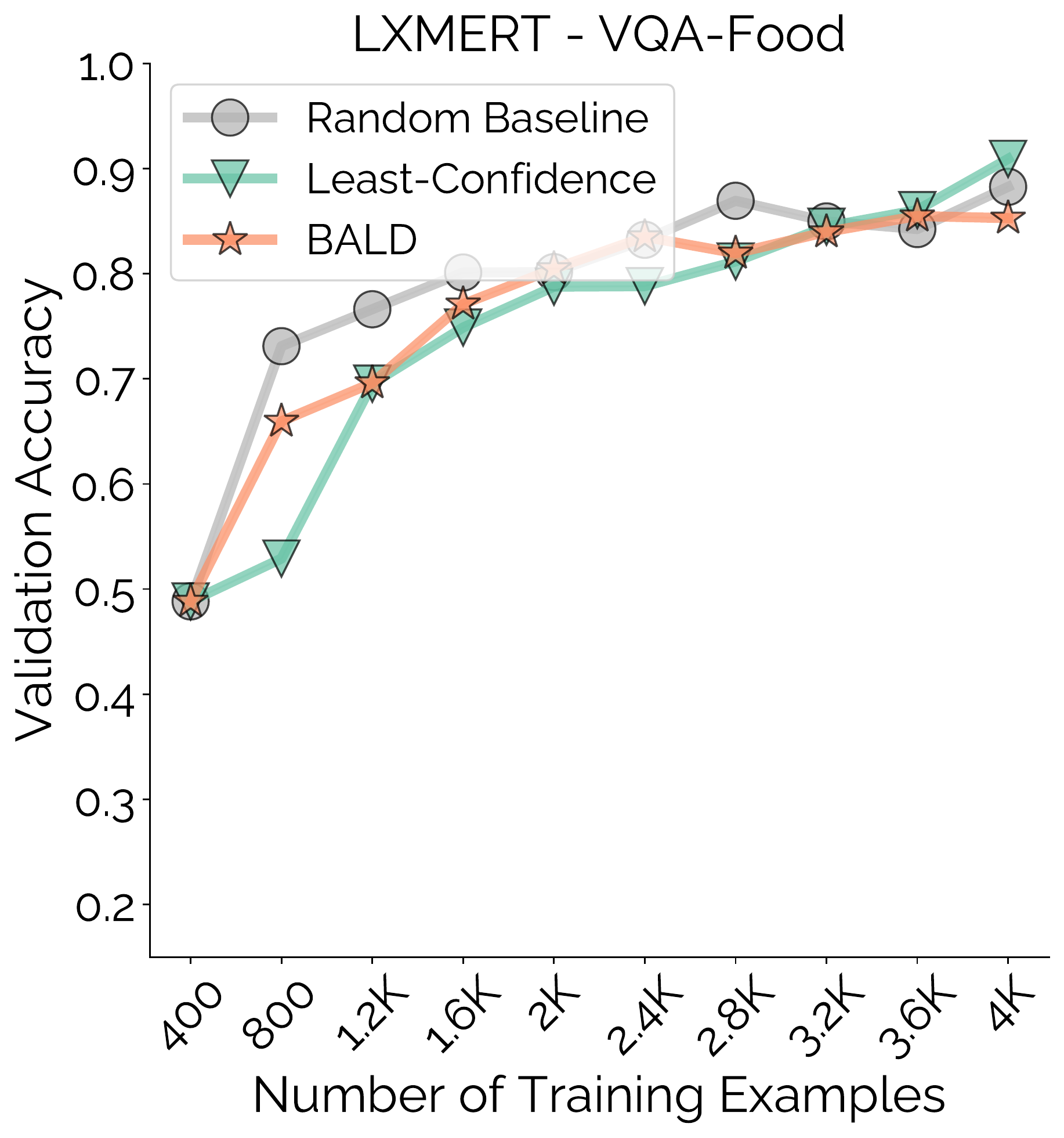}
    \end{subfigure}
    \vspace*{-5pt}
    \caption{Results for the representative active learning methods on VQA-Food, a simplified VQA dataset similar to VQA-Food, across LSTM-CNN, BUTD, and LXMERT.}
    \label{sfig:vqa-food-p10}
\end{figure*}

Figure \ref{sfig:vqa-food-p10} shows results on VQA-Food with the LSTM-CNN, BUTD, and LXMERT models, with a seed set comprised of 10\% of the total pool. The results are mostly similar to those reported in the paper; strategies track or underperform random sampling, with the exception of Least-Confidence for the LSTM-CNN model -- however, this is the sole exception, and the LSTM-CNN has the highest training variance of all the models we try.

\subsection{Logistic Regression (ResNet-101)}

\begin{figure*}
    \centering
    \begin{subfigure}[b]{0.31\textwidth}
        \centering
        \includegraphics[width=\textwidth, height=0.2\textheight]{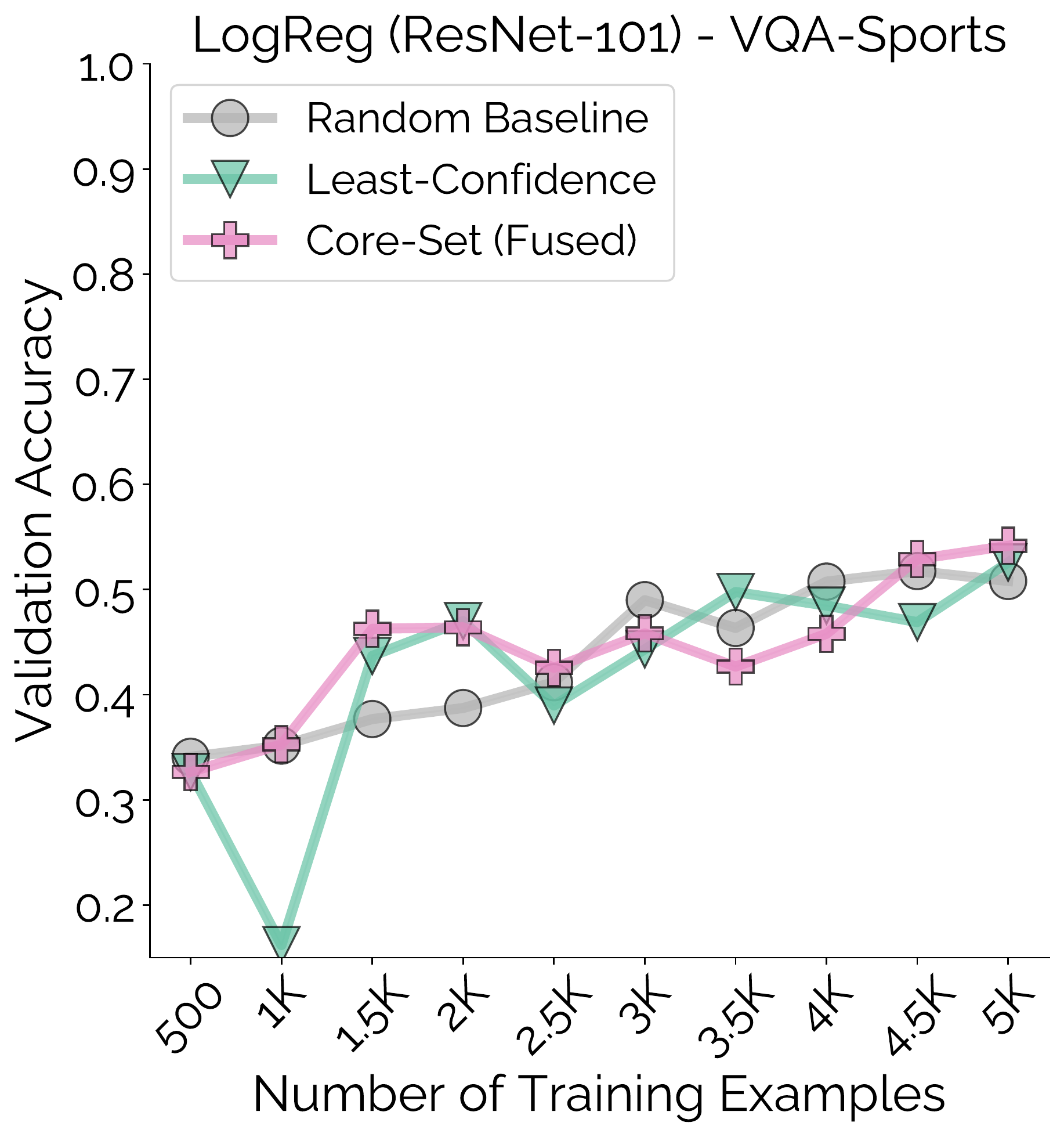}
    \end{subfigure}
    \hfill
    \begin{subfigure}[b]{0.31\textwidth}
        \centering
        \includegraphics[width=\textwidth, height=0.2\textheight]{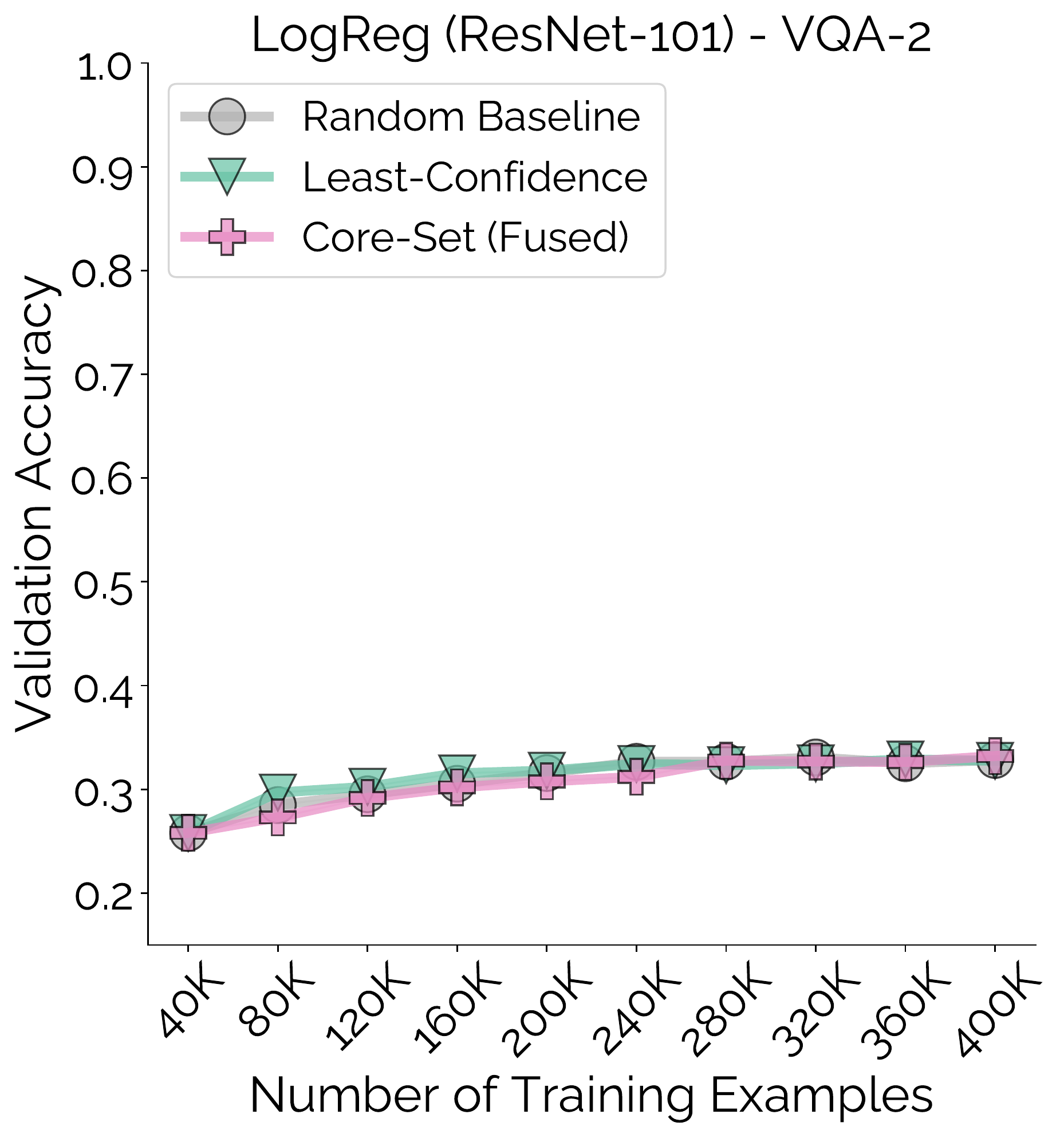}
    \end{subfigure}
    \hfill
    \begin{subfigure}[b]{0.31\textwidth}
        \centering
        \includegraphics[width=\textwidth, height=0.2\textheight]{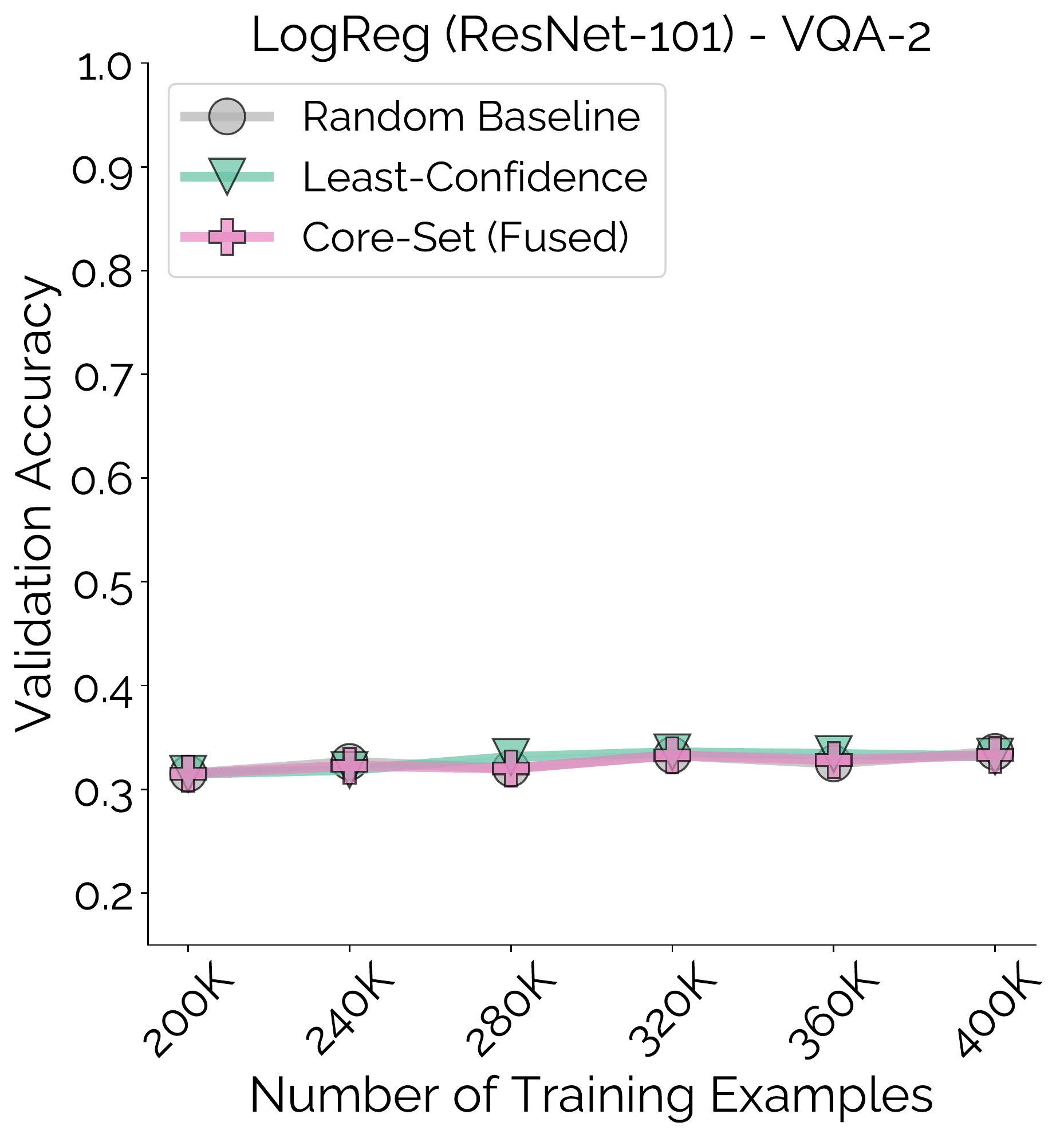}
    \end{subfigure}
    \vspace*{-5pt}
    \caption{Active learning results using the Logistic Regression (ResNet-101) model on VQA-Sports (10\% seed set), and VQA-2 (10\% and 50\% seed set). Most strategies either track or underperform random acquisition.}    
    \label{sfig:glreg-p10}
\end{figure*}

Figure \ref{sfig:glreg-p10} shows active learning results for the LogReg (ResNet-101) model on VQA-Sports (seed set = 10\%), and VQA-2 (seed set = 10\%, 50\%). Results are similar to those reported in the paper, with active learning failing to outperform random acqusition.

\subsection{Logistic Regression (Faster R-CNN)}

\begin{figure*}
    \centering
    \begin{subfigure}[b]{0.31\textwidth}
        \centering
        \includegraphics[width=\textwidth, height=0.2\textheight]{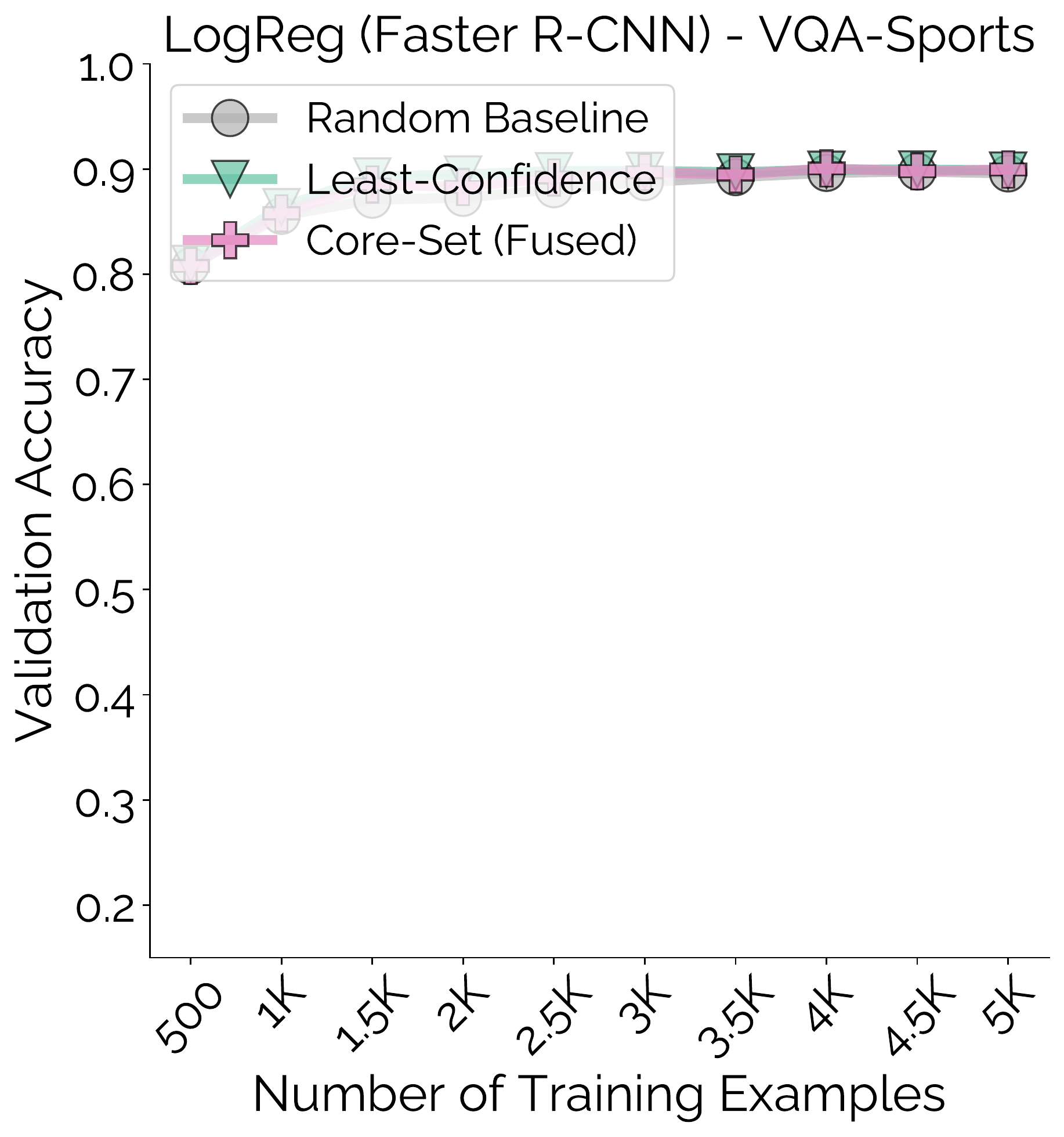}
    \end{subfigure}
    \hfill
    \begin{subfigure}[b]{0.31\textwidth}
        \centering
        \includegraphics[width=\textwidth, height=0.2\textheight]{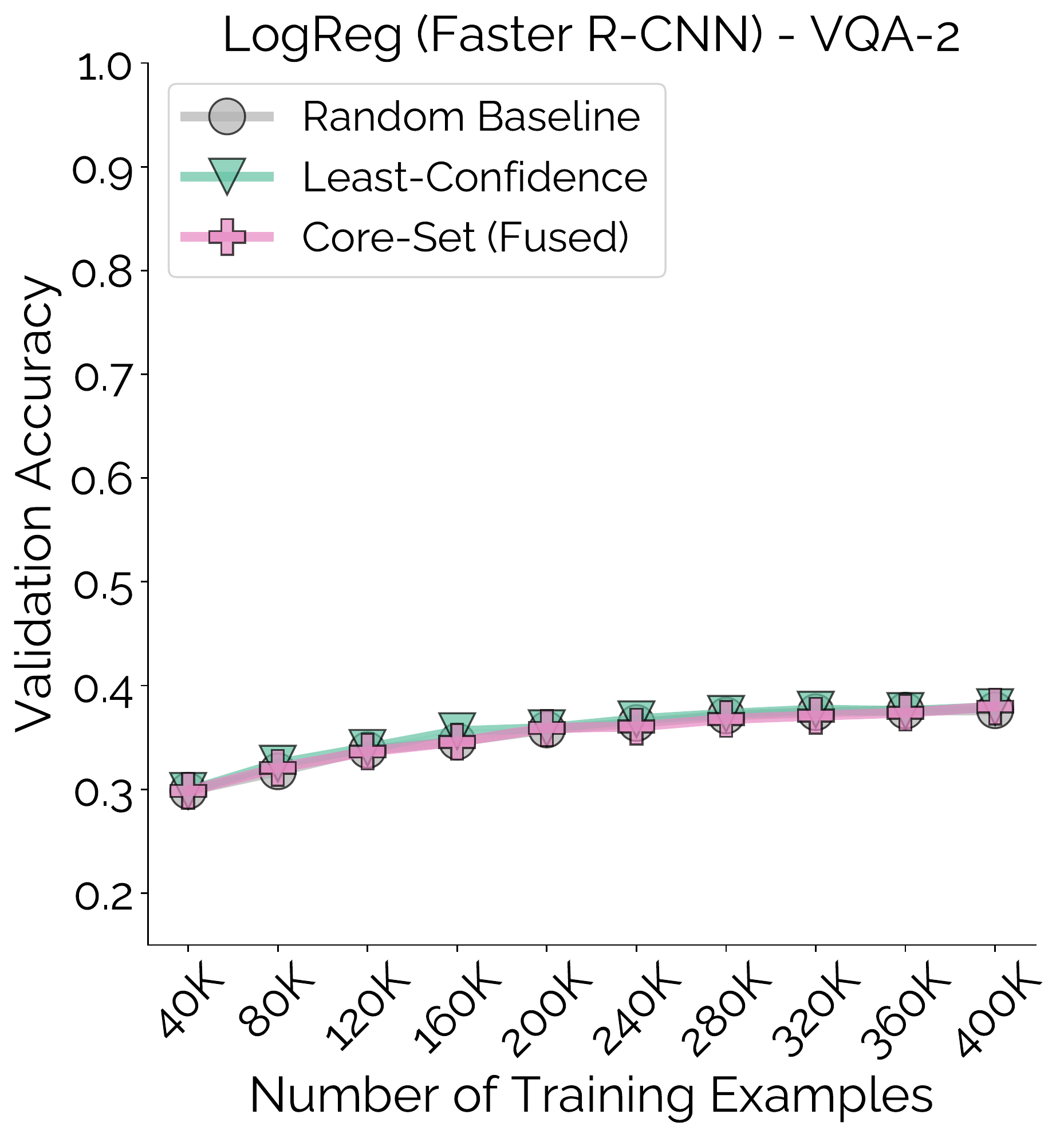}
    \end{subfigure}
    \hfill
    \begin{subfigure}[b]{0.31\textwidth}
        \centering
        \includegraphics[width=\textwidth, height=0.2\textheight]{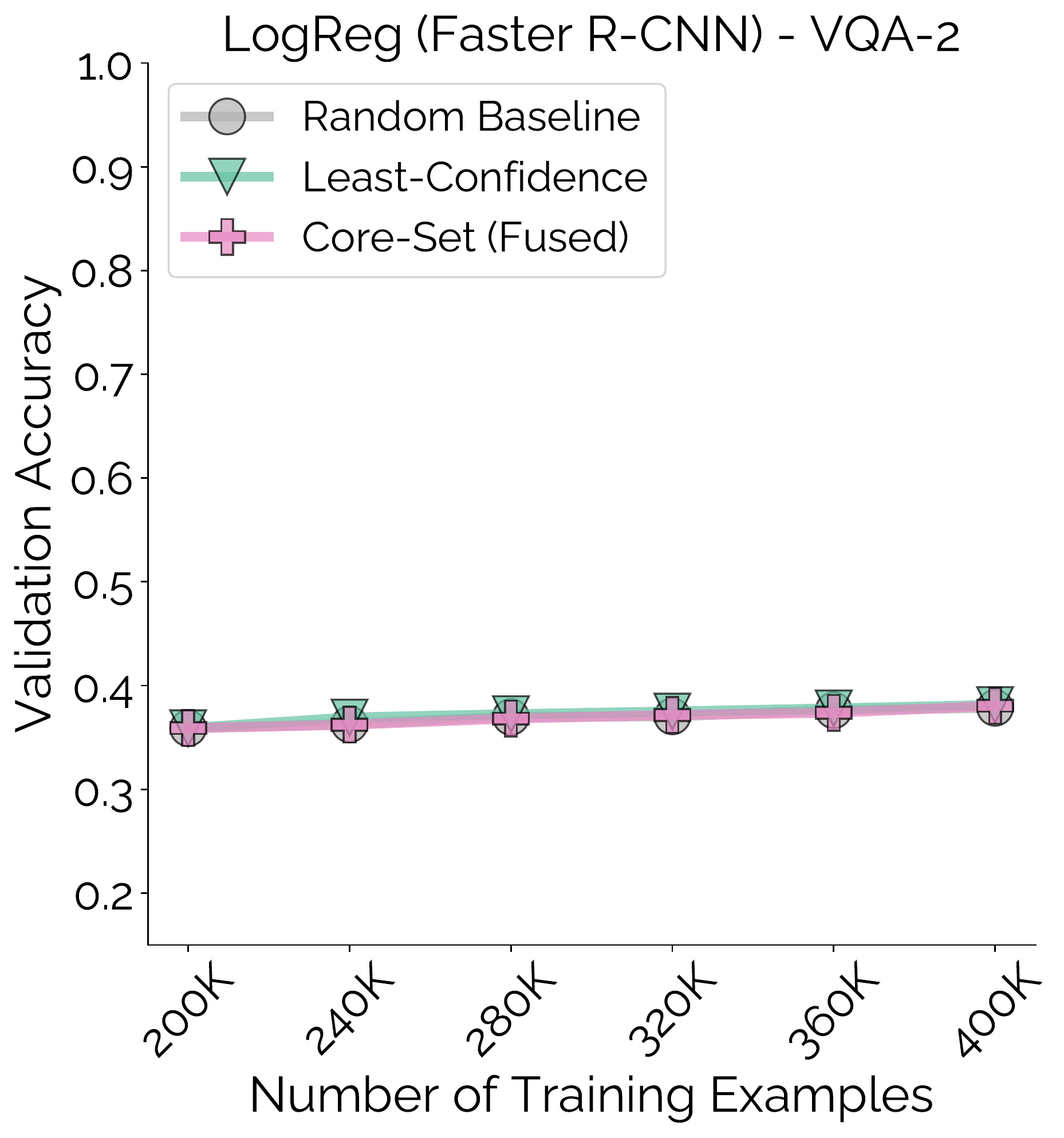}
    \end{subfigure}
    \vspace*{-5pt}
    \caption{Active learning results using the Logistic Regression (Faster R-CNN) model on VQA-Sports (10\% seed set), and VQA-2 (10\% and 50\% seed set). While the Faster R-CNN representation leads to better validation accuracies, active learning performance remains consistent.} 
    \label{sfig:olreg-p10}
\end{figure*}

Figure \ref{sfig:olreg-p10} presents the same set of experiments as the prior section, except with the LogReg (Faster R-CNN) model. While the object-based Faster R-CNN representation enables much higher performance than the ResNet-101 representation, active learning results are consistent with those reported in the paper.

\subsection{Other Acquisition Strategies}

\begin{figure*}
    \centering
    \begin{subfigure}[b]{0.31\textwidth}
        \centering
        \includegraphics[width=\textwidth, height=0.2\textheight]{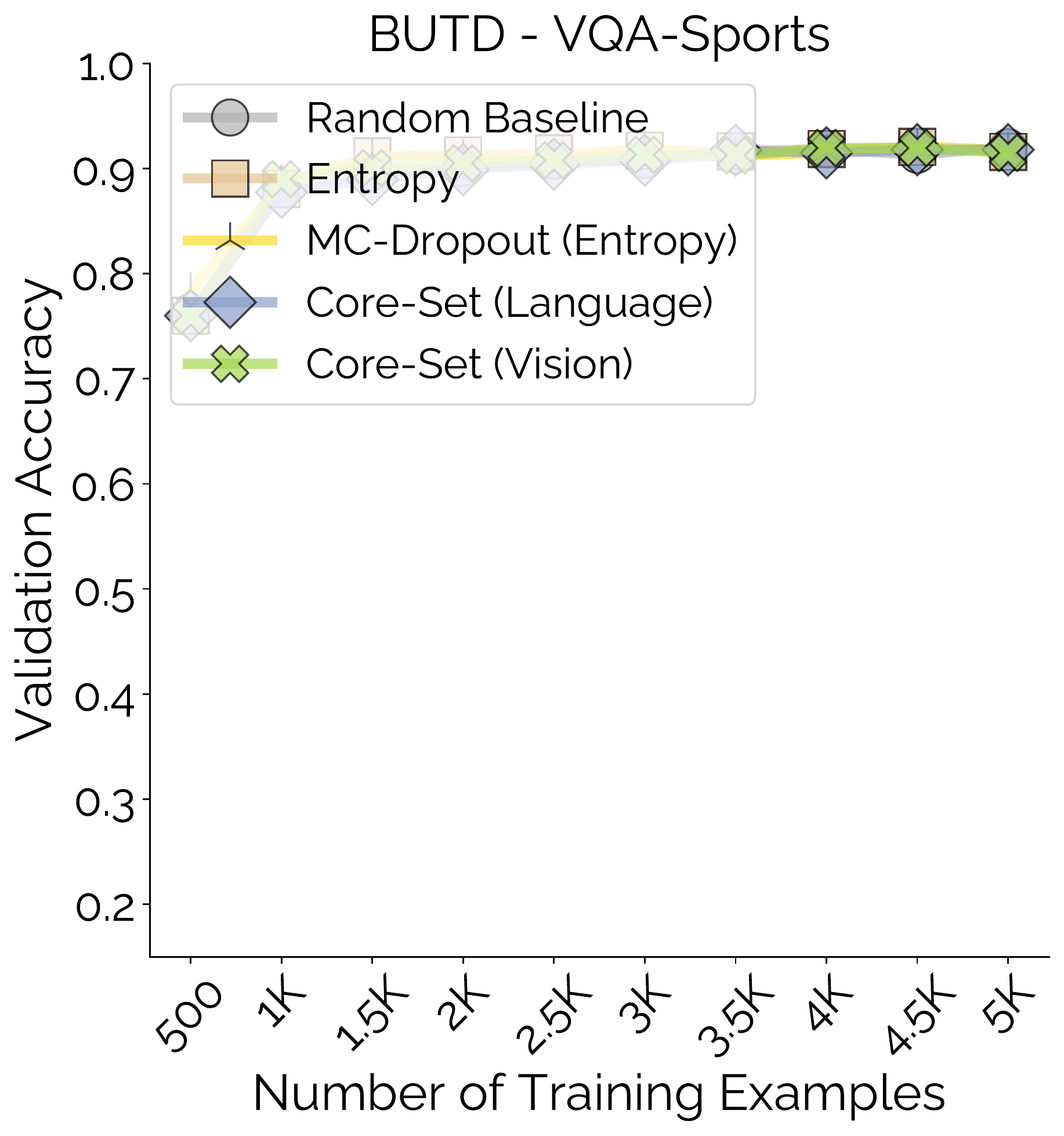}
    \end{subfigure}
    \hfill
    \begin{subfigure}[b]{0.31\textwidth}
        \centering
        \includegraphics[width=\textwidth, height=0.2\textheight]{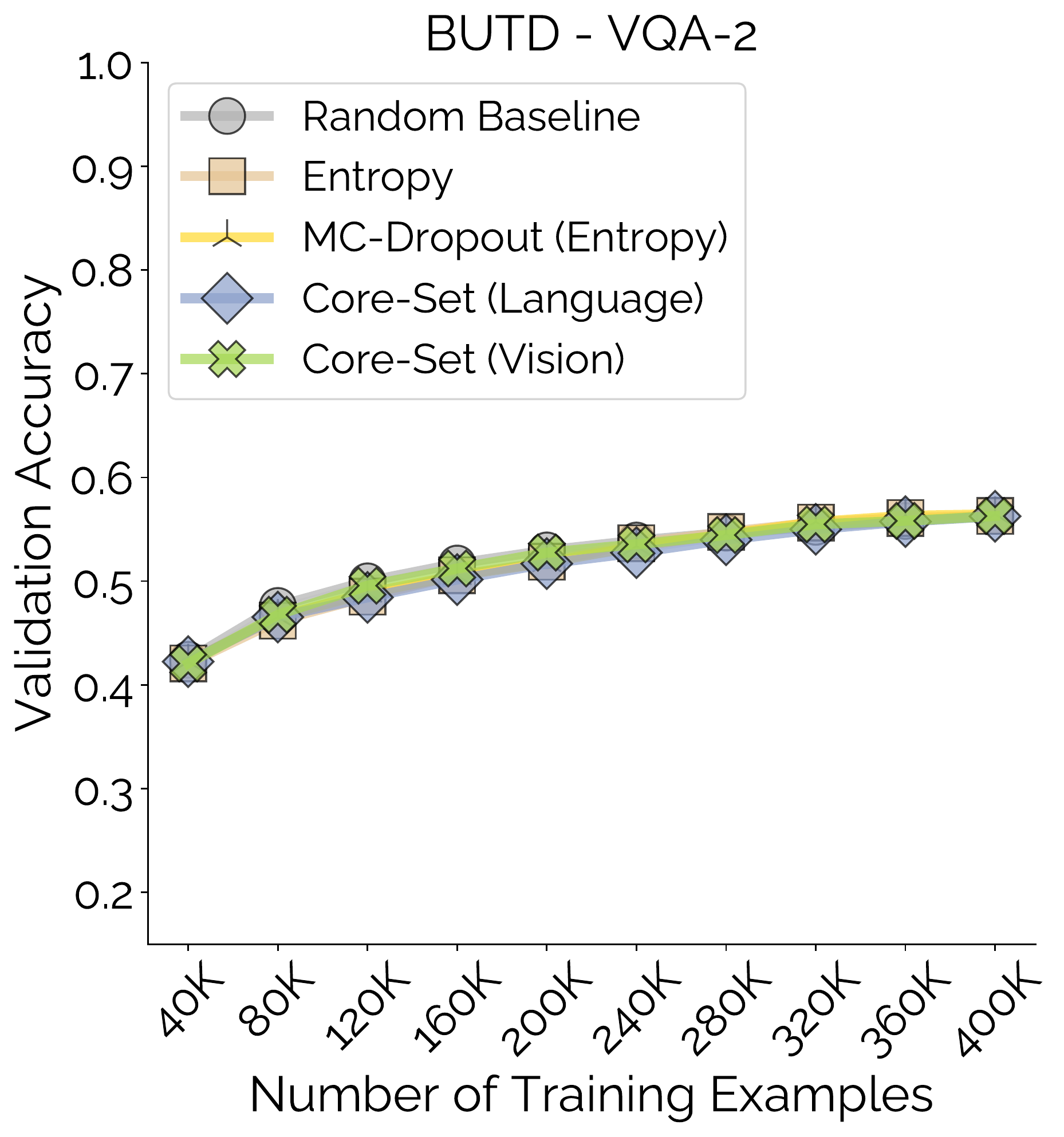}
    \end{subfigure}
    \hfill
    \begin{subfigure}[b]{0.31\textwidth}
        \centering
        \includegraphics[width=\textwidth, height=0.2\textheight]{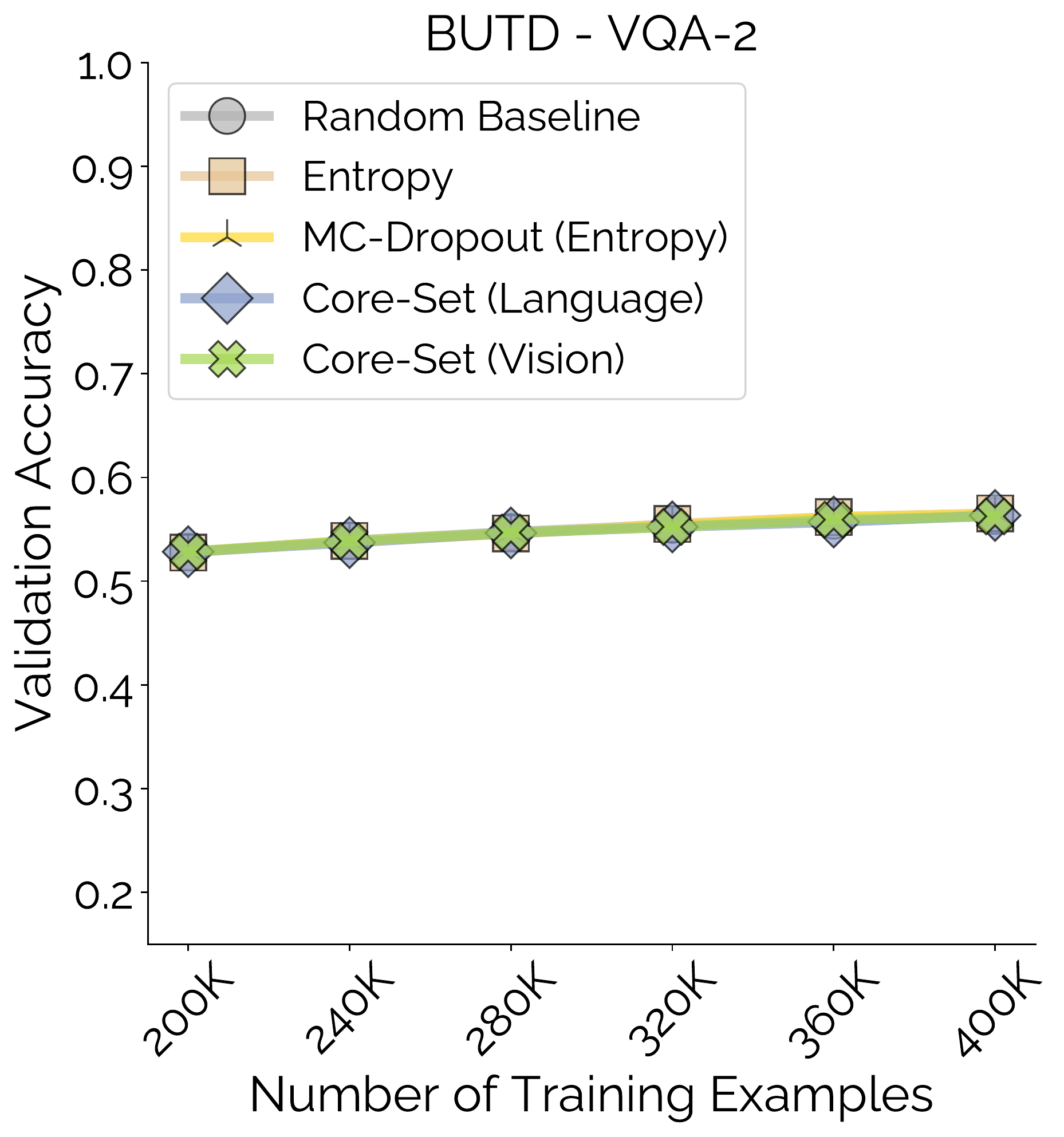}
    \end{subfigure}
    \vspace*{-5pt}
    \caption{Results with the BUTD on VQA-Sports, VQA-2 and GQA using the alternative 4 acquisition strategies not included in the main body of the paper. Unsurprisingly, results are consistent with those reported in the paper.}    
    \label{sfig:alt-strats-p10}
\end{figure*}

Figure \ref{sfig:alt-strats-p10} presents results for the four other active learning strategies we implement -- Entropy, Monte Carlo Dropout w/ Entropy, Core-Set (Language), and Core-Set (Vision) -- for the BUTD model. Results are across VQA-Sports (seed set = 10\%), and VQA-2 (seed set = 10\%, 50\%) -- despite the unique features of each strategy, the trends remain consistent with those in the paper.

\section{Dataset Maps \& Acquisitions}

\begin{figure*}
    \centering
    \begin{subfigure}[b]{0.31\textwidth}
        \centering
        \includegraphics[width=\textwidth, height=0.2\textheight]{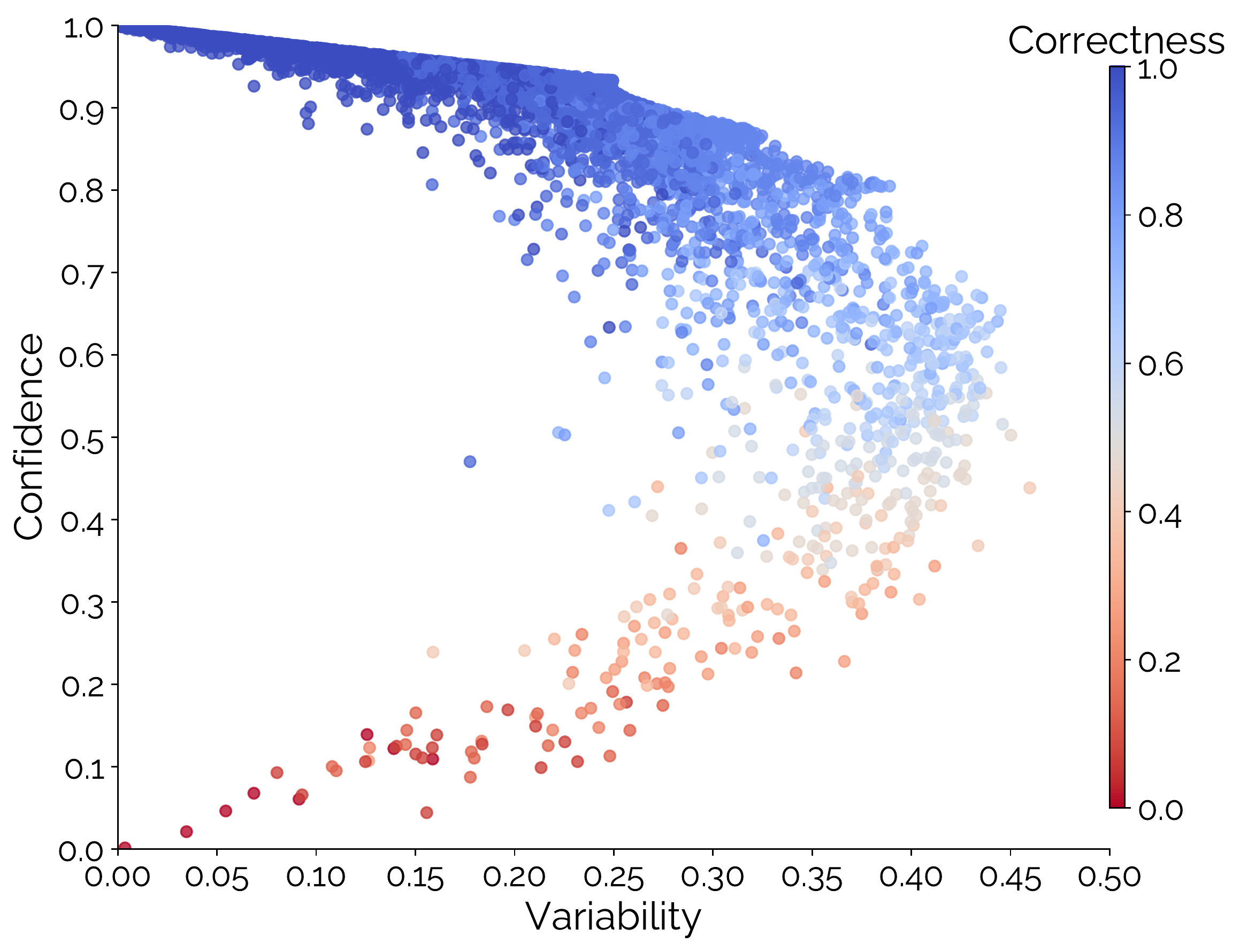}
    \end{subfigure}
    \hfill
    \begin{subfigure}[b]{0.31\textwidth}
        \centering
        \includegraphics[width=\textwidth, height=0.2\textheight]{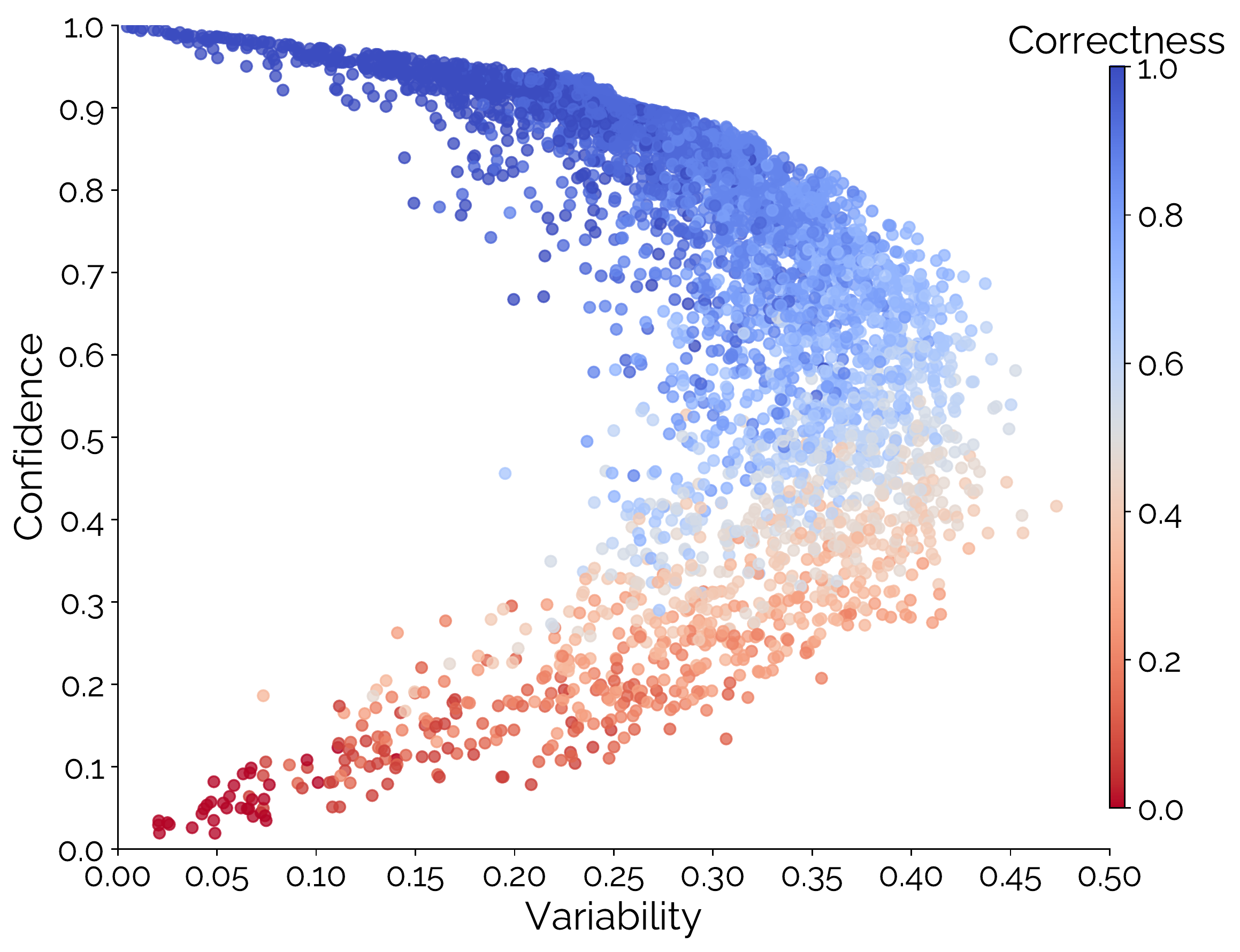}
    \end{subfigure}
    \hfill
    \begin{subfigure}[b]{0.31\textwidth}
        \centering
        \includegraphics[width=\textwidth, height=0.2\textheight]{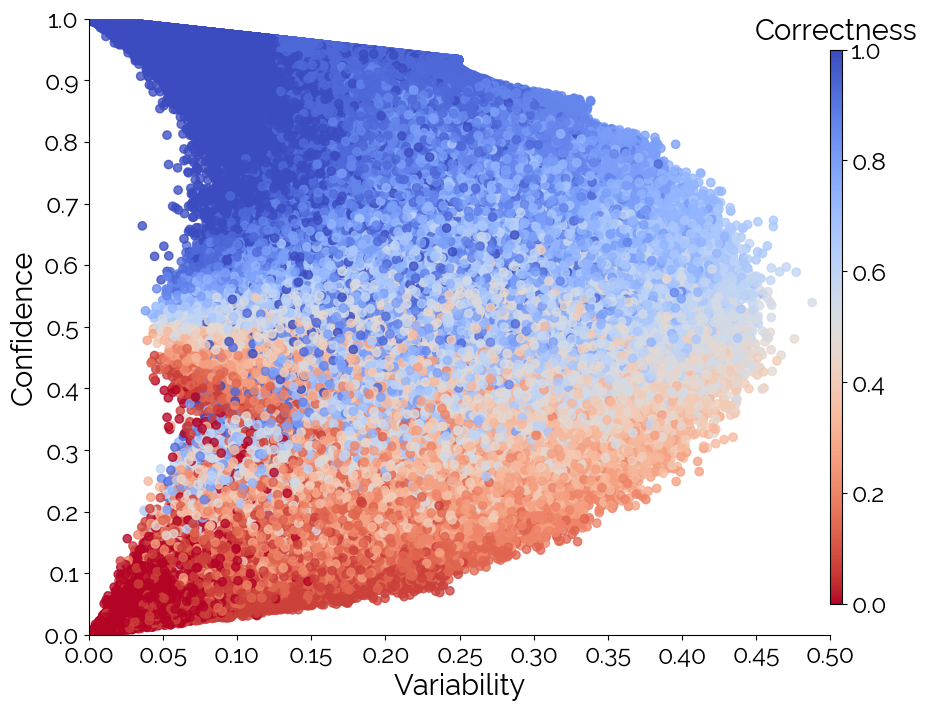}
    \end{subfigure}
    \vspace*{-5pt}
    \caption{Dataset Maps for the Bottom-Up Top-Down Attention model on VQA-Sports, VQA-Food, and GQA respectively. Note that VQA-Sports and VQA-Food have fewer ``hard-to-learn'' examples.}
    \label{sfig:butd-maps}
\end{figure*}

\begin{figure*}
    \centering
    \begin{subfigure}[b]{0.31\textwidth}
        \centering
        \includegraphics[width=\textwidth, height=0.2\textheight]{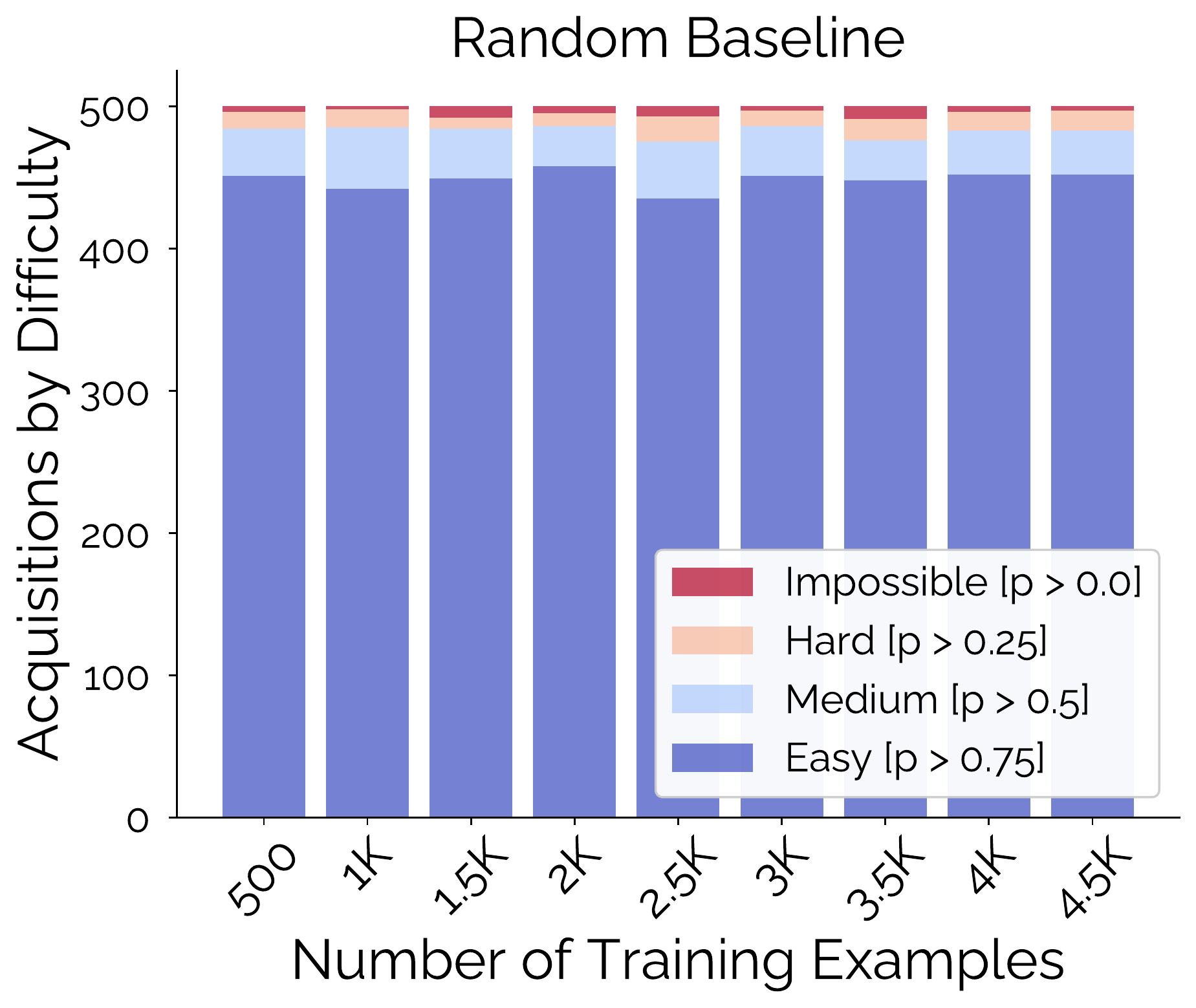}
    \end{subfigure}
    \hfill
    \begin{subfigure}[b]{0.31\textwidth}
        \centering
        \includegraphics[width=\textwidth, height=0.2\textheight]{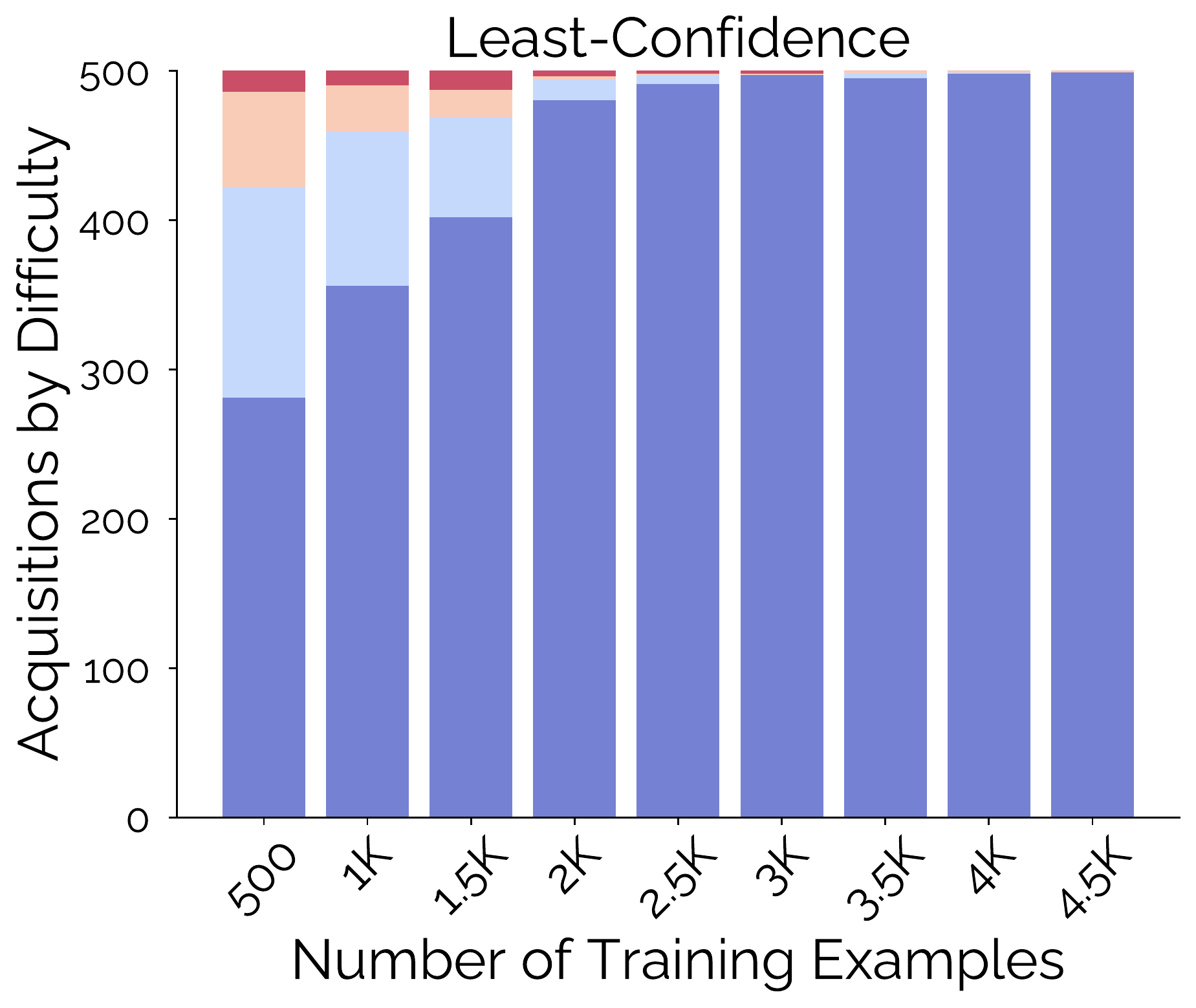}
    \end{subfigure}
    \hfill
    \begin{subfigure}[b]{0.31\textwidth}
        \centering
        \includegraphics[width=\textwidth, height=0.2\textheight]{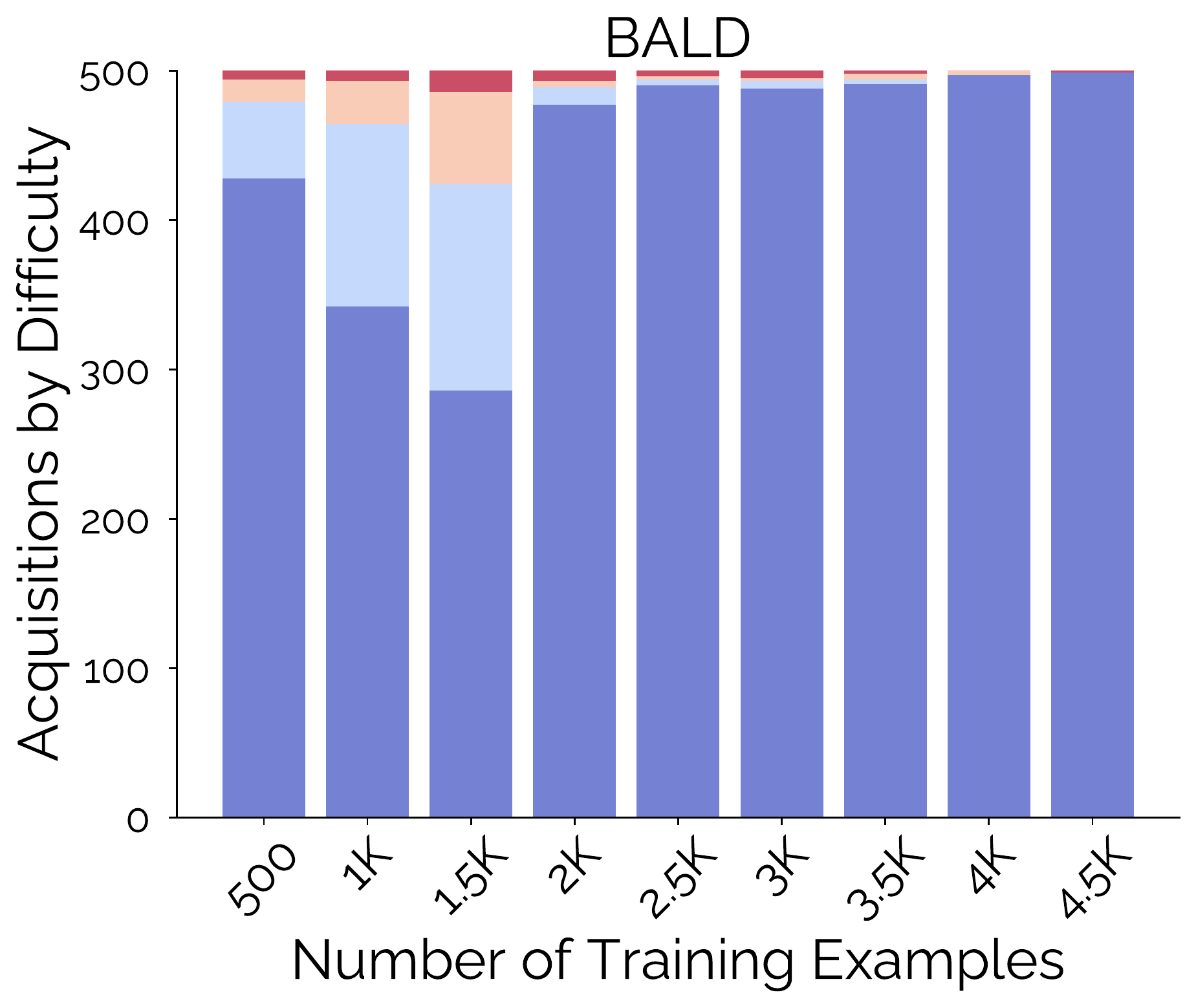}
    \end{subfigure}
    \vspace*{-5pt}
    \caption{Acquisitions with the BUTD Model on VQA-Sports. The dataset has fewer ``hard-to-learn'' examples, but active learning strategies pick the medium--hard examples, which still negatively impact performance.}    
    \label{sfig:sports-acquisitions}
\end{figure*}

\begin{figure*}
    \centering
    \begin{subfigure}[b]{0.31\textwidth}
        \centering
        \includegraphics[width=\textwidth, height=0.2\textheight]{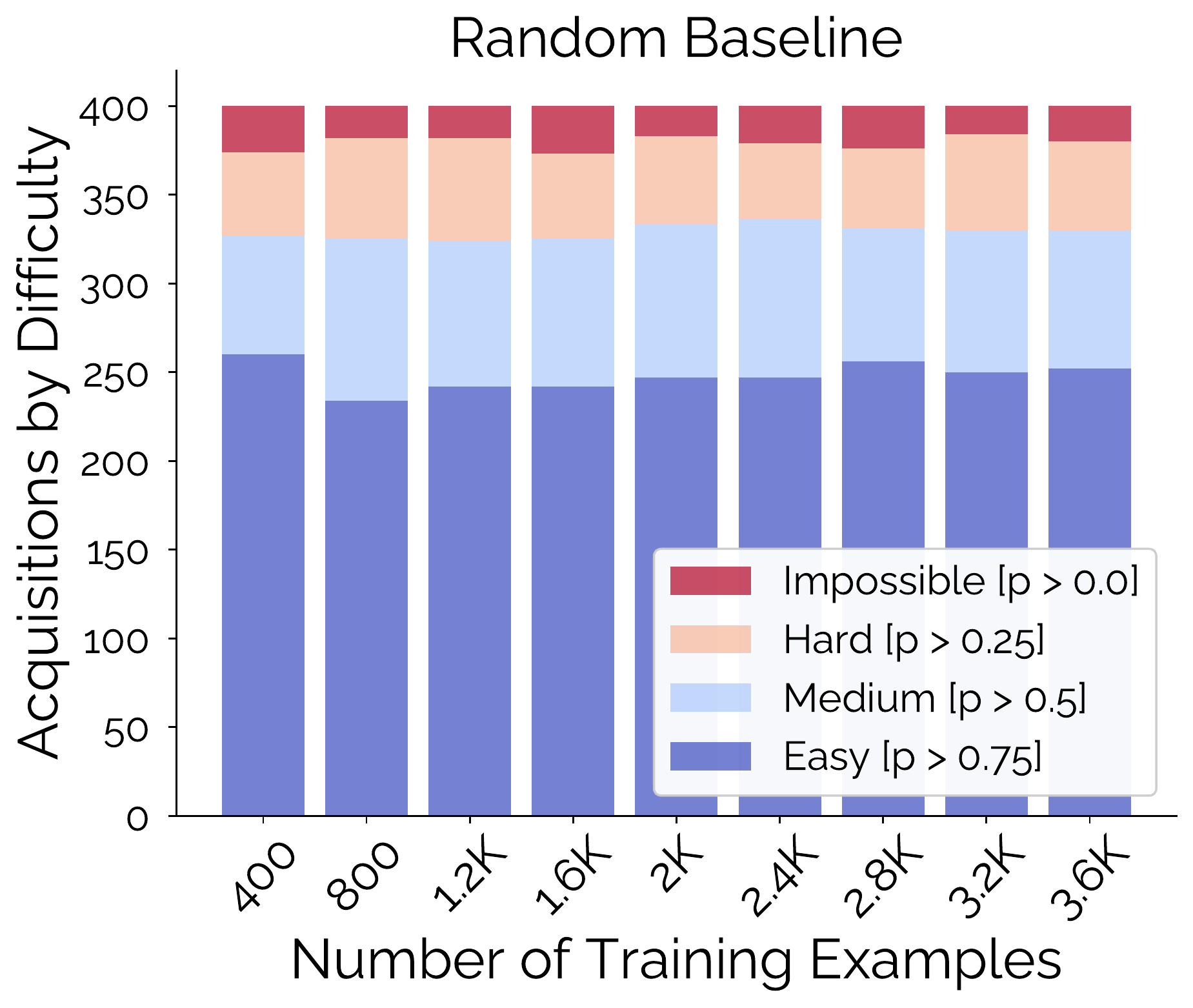}
    \end{subfigure}
    \hfill
    \begin{subfigure}[b]{0.31\textwidth}
        \centering
        \includegraphics[width=\textwidth, height=0.2\textheight]{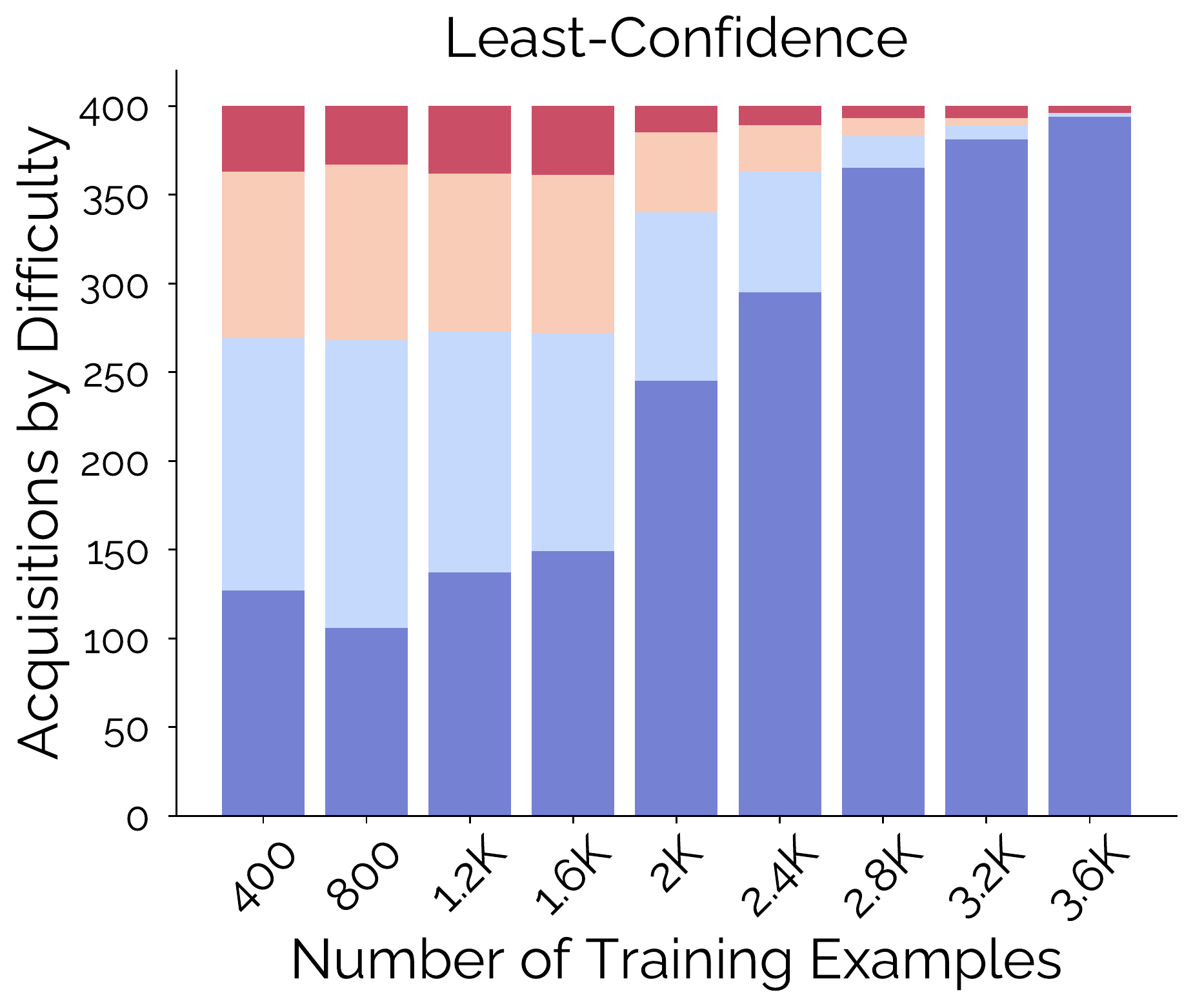}
    \end{subfigure}
    \hfill
    \begin{subfigure}[b]{0.31\textwidth}
        \centering
        \includegraphics[width=\textwidth, height=0.2\textheight]{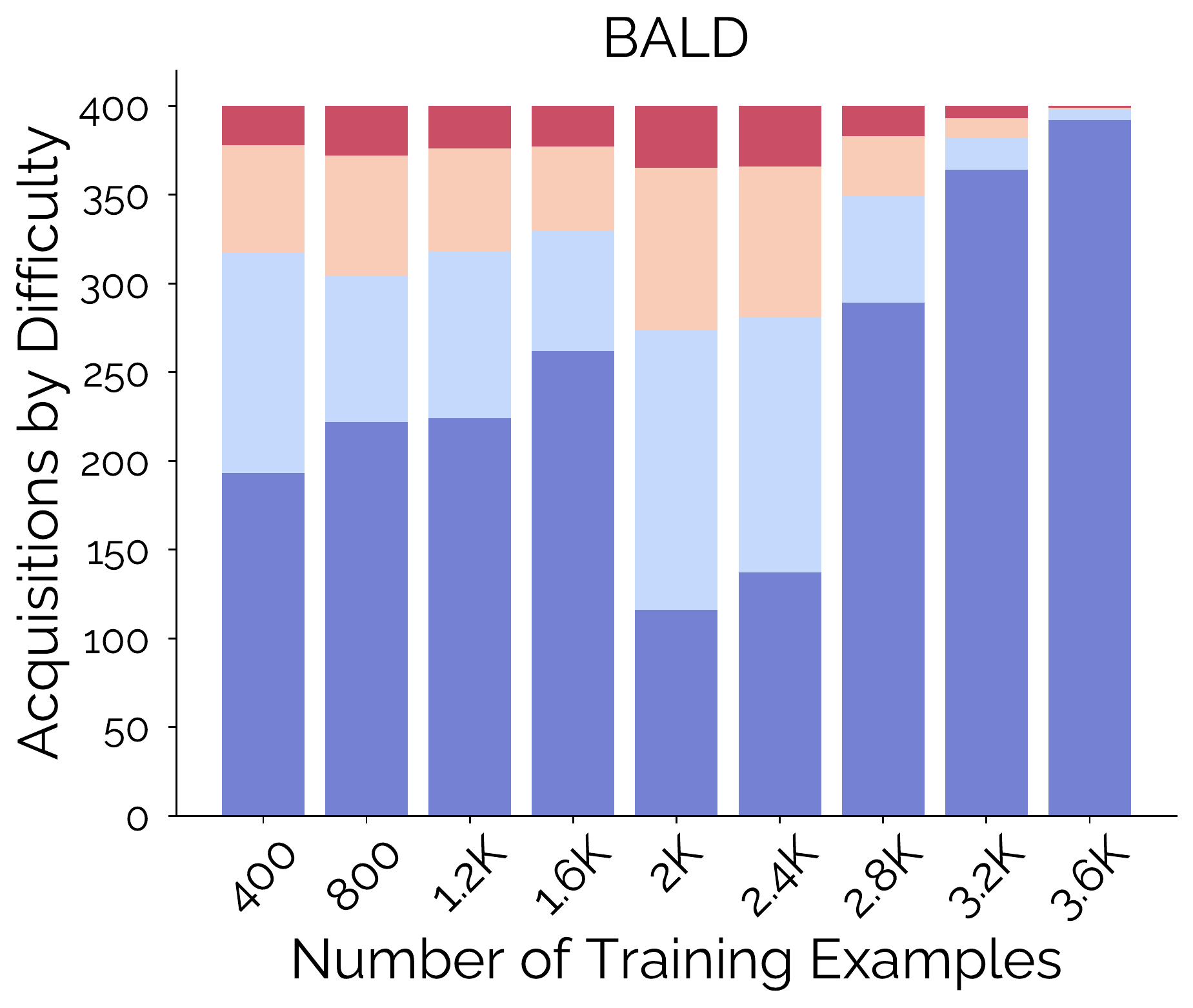}
    \end{subfigure}
    \vspace*{-5pt}
    \caption{Acquisitions with the BUTD Model on VQA-Food. Despite the sparsity of hard examples, active learning strategies still tend towards them. BALD is high-variance, selecting examples all over the map.} 
    \label{sfig:food-acquisitions}
\end{figure*}

\begin{figure*}
    \centering
    \begin{subfigure}[b]{0.31\textwidth}
        \centering
        \includegraphics[width=\textwidth, height=0.2\textheight]{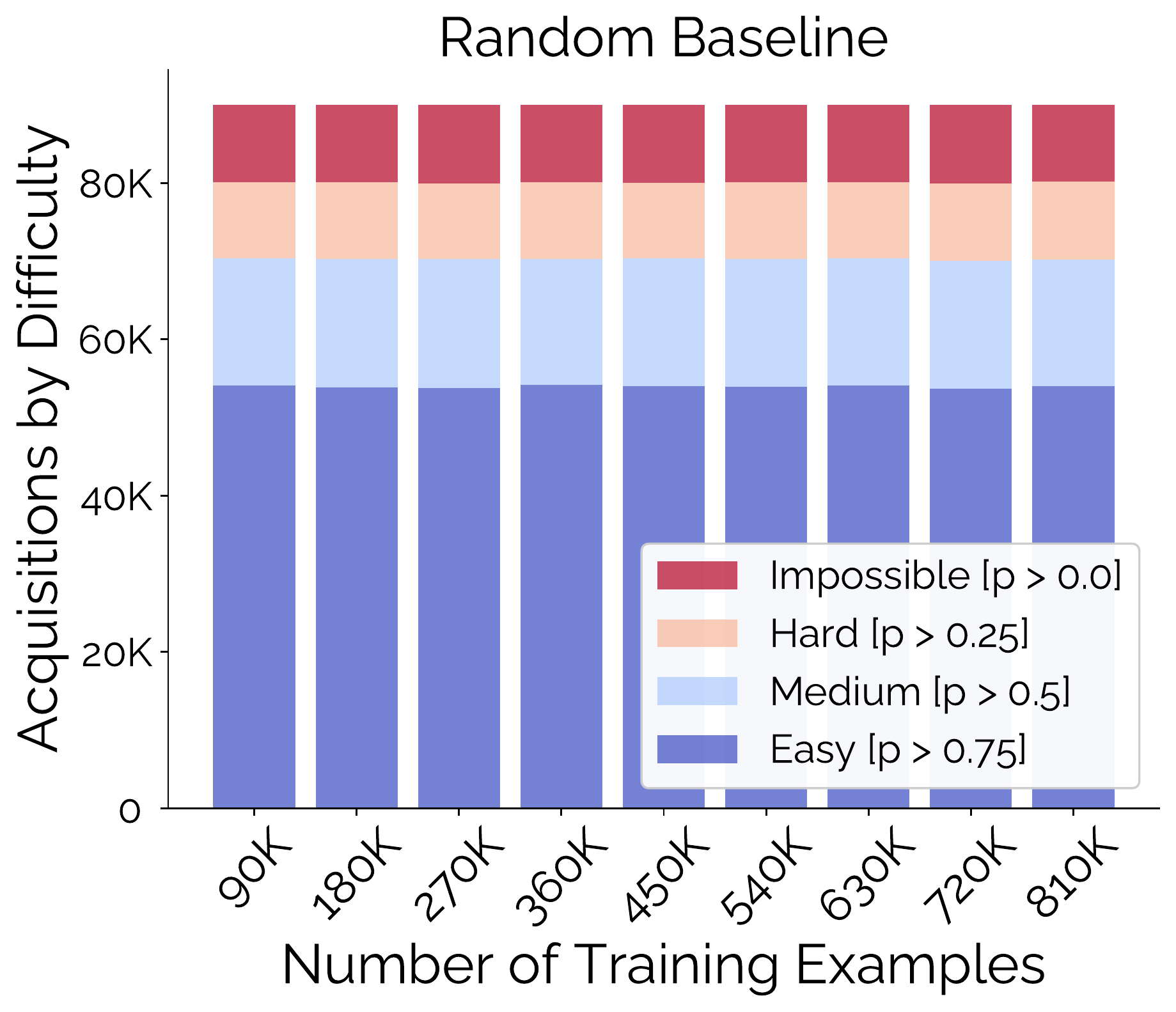}
    \end{subfigure}
    \hfill
    \begin{subfigure}[b]{0.31\textwidth}
        \centering
        \includegraphics[width=\textwidth, height=0.2\textheight]{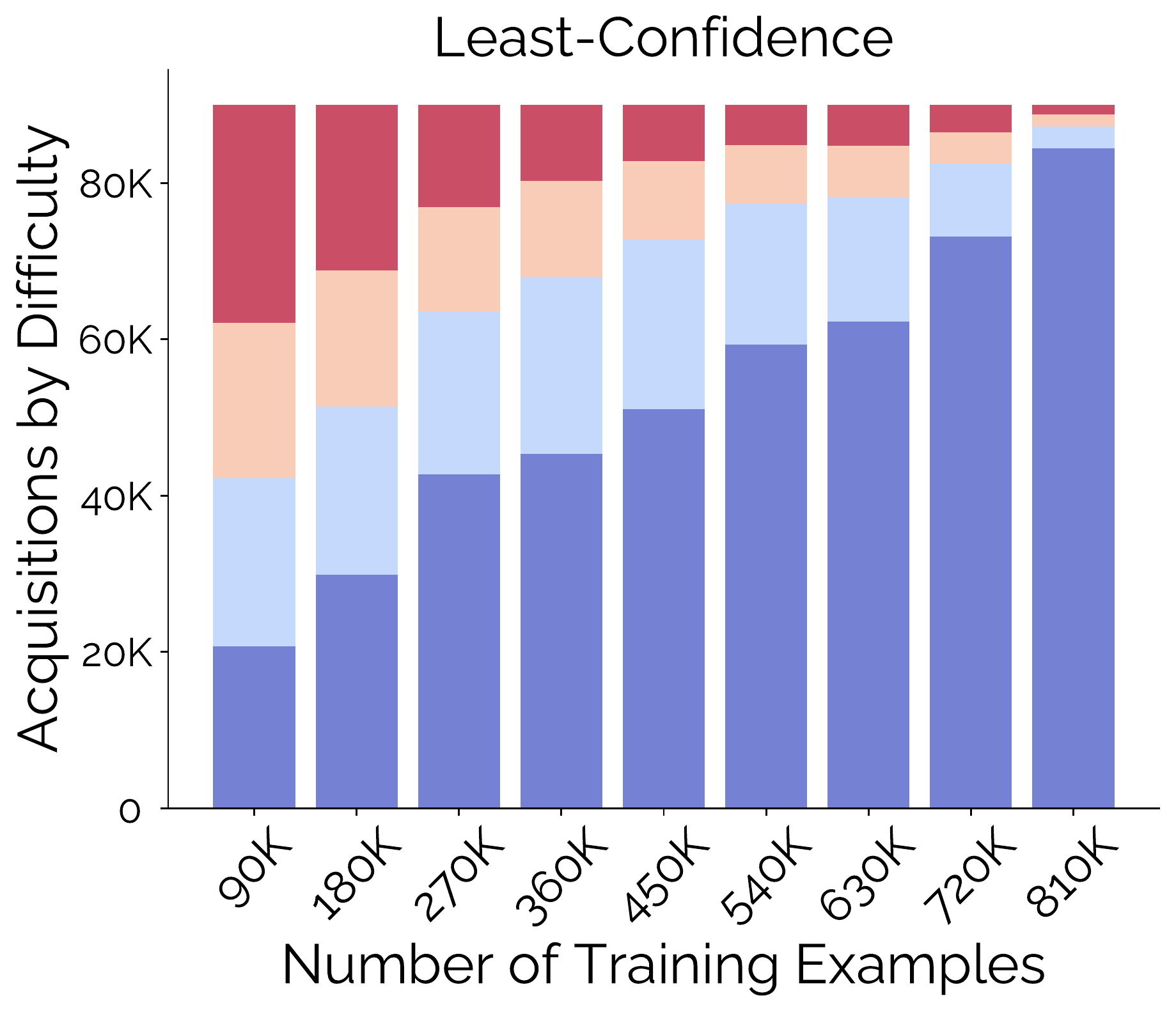}
    \end{subfigure}
    \hfill
    \begin{subfigure}[b]{0.31\textwidth}
        \centering
        \includegraphics[width=\textwidth, height=0.2\textheight]{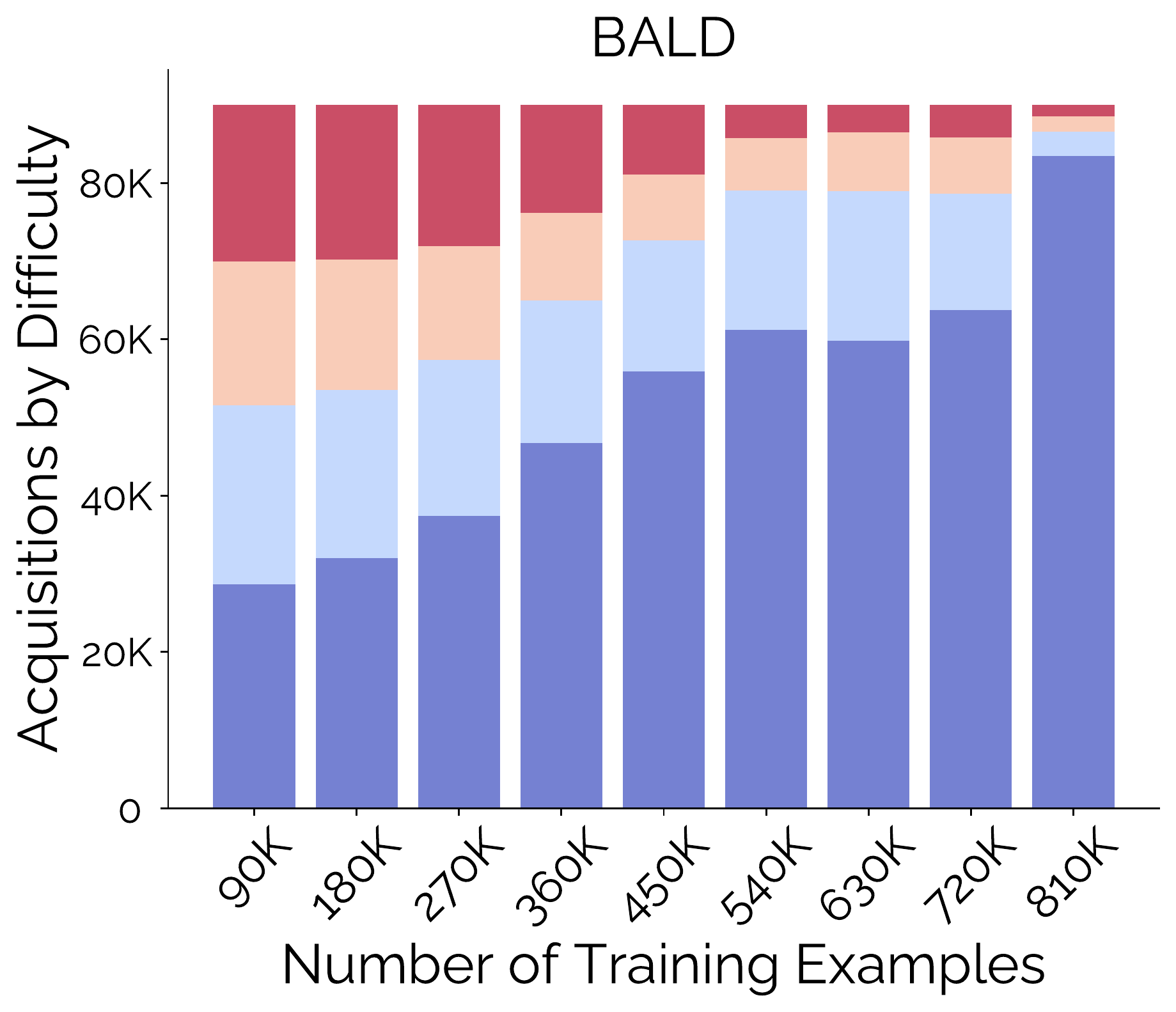}
    \end{subfigure}
    \vspace*{-5pt}
    \caption{Acquisitions with the BUTD Model on the full GQA dataset. Given that the map for GQA is similar to the map for VQA-2, it is not surprising that the active learning acquisitions follow a similar trend, preferring to select ``hard-to-learn'' examples.}    
    \label{sfig:gqa-acquisitions}
\end{figure*}

To provide further context around active learning acquisitions across datasets, Figures \ref{sfig:butd-maps}--\ref{sfig:gqa-acquisitions} present Dataset Maps and acquisitions for the BUTD Model across VQA-Sports, VQA-Food, and GQA respectively. Interesting to note is that while VQA-Sports and VQA-Food are generally easier, with fewer ``hard-to-learn'' examples, active learning still has a bias for picking those examples. For GQA, our earlier analysis is confirmed; active learning is picking the collective outliers populating the bottom half of the Dataset Map.
\setlength{\belowcaptionskip}{-3pt}

\setlength{\belowcaptionskip}{-3pt}

\end{document}